%% file: main.tex
\documentclass{article} % For LaTeX2e
\usepackage[preprint]{neurips_2025}

\input{math_commands.tex}

\usepackage[utf8]{inputenc} % allow utf-8 input
\usepackage[T1]{fontenc}    % use 8-bit T1 fonts
\usepackage{hyperref}       % hyperlinks
\usepackage{url}            % simple URL typesetting
\usepackage{booktabs}       % professional-quality tables
\usepackage{amsfonts}       % blackboard math symbols
\usepackage{nicefrac}       % compact symbols for 1/2, etc.
\usepackage{microtype}      % microtypography
\usepackage{xcolor}         % colors

\usepackage{multirow}
\usepackage{amsmath}
\usepackage{graphicx}
\usepackage{algorithm}
\usepackage{algorithmic}
\usepackage{xr}

\title{DiffER$^2$: Diffusion Ensembles for Retrosynthesis Prediction with SMILES Adapted Particle Guidance}

\author{%
  Sean F.~Current \\
  Department of Computer Science \\
  The Ohio State University \\
  Columbus, OH 43202 \\
  \texttt{current.33@osu.edu} \\
  \And
  Daniel Adu-Ampratwum \\
  College of Pharmacy \\
  The Ohio State University \\
  Columbus, OH 43202 \\
  \texttt{adu-ampratwum.1@osu.edu} \\
  \And
  Xia Ning \\
  Department of Computer Science \\
  Department of Biomedical Informatics \\
  The Ohio State University \\
  Columbus, OH 43202 \\
  \texttt{ning.104@osu.edu} \\
  \And
  Srinivasan Parthasarathy \\
  Department of Computer Science \\
  The Ohio State University \\
  Columbus, OH 43202 \\
  \texttt{srini@cse.ohio-state.edu} \\
 }

\begin{document}

\maketitle

\begin{abstract}
 Computational retrosynthesis prediction has the potential to reduce development time for newly discovered drugs by automatically generating potential reactions for a target product. Data-driven approaches commonly treat this as a sequence-to-sequence generation task on SMILES strings. In this work, we construct ensembles of discrete-time and continuous-time diffusion models for molecular generation and incorporate guidance mechanisms for improved output diversity. We propose an adaptation of particle guidance to SMILES sequence generation which significantly improves the number of unique molecules generated by diffusion ensembles while increasing top-k accuracy. These results further expand the efficacy of discrete diffusion for SMILES generation, and our empirical analyses offer new insights into the capabilities of diffusion models for chemical retrosynthesis.
\end{abstract}

\section{Introduction}

The goal of retrosynthesis is to construct reactant molecules that can be used to synthesize a target product. Retrosynthesis prediction constitutes a pivotal step in the drug manufacturing pipeline. Repeatedly applying single-step retrosynthesis models to a given problem can result in full synthetic pathway discovery \citep{segler2018planning, shen2021automation}, further supporting pharmaceutical advancement for new drug modalities. However, even single-step retrosynthesis prediction is difficult, in part due to the diversity of reactant sets that might map to a single product. While forward synthesis prediction is primarily considered a surjective process with a deterministic output, retrosynthesis prediction is in many ways the opposite, with many possible outputs resulting from a single input. As a result, researchers have taken a particular interest in deep learning methods that hasten the process of retrosynthesis discovery by quickly and efficiently predicting multiple possible reactant sets from a given product molecule.

Numerous methodologies have been developed, from template-based methods that use mined reaction mechanisms to predict reactants to template-free approaches that directly predict string or graph representations of reactant molecules conditioned on the target product. The most successful of these template-free approaches use traditional transformer encoder-decoder architectures on SMILES \citep{weininger1988smiles} string representations of the molecules, effectively treating the problem as a machine translation task \citep{zhong2022root}. A recent approach, DiffER \citep{current2024differ}, explored the usage of discrete diffusion models for the retrosynthesis task and provided evidence that an ensemble of discrete-time diffusion models could be a practical approach for sampling from the posterior distribution of reactants. Differ outperformed the state-of-the-art R-SMILES \citep{zhong2022root} at top-1 accuracy on the commonly used USPTO-50k dataset \citep{Lowe2017}, but suffered from low output diversity due to the nature of posterior sampling.

\begin{figure}
    \centering
    \includegraphics[width=0.95\linewidth]{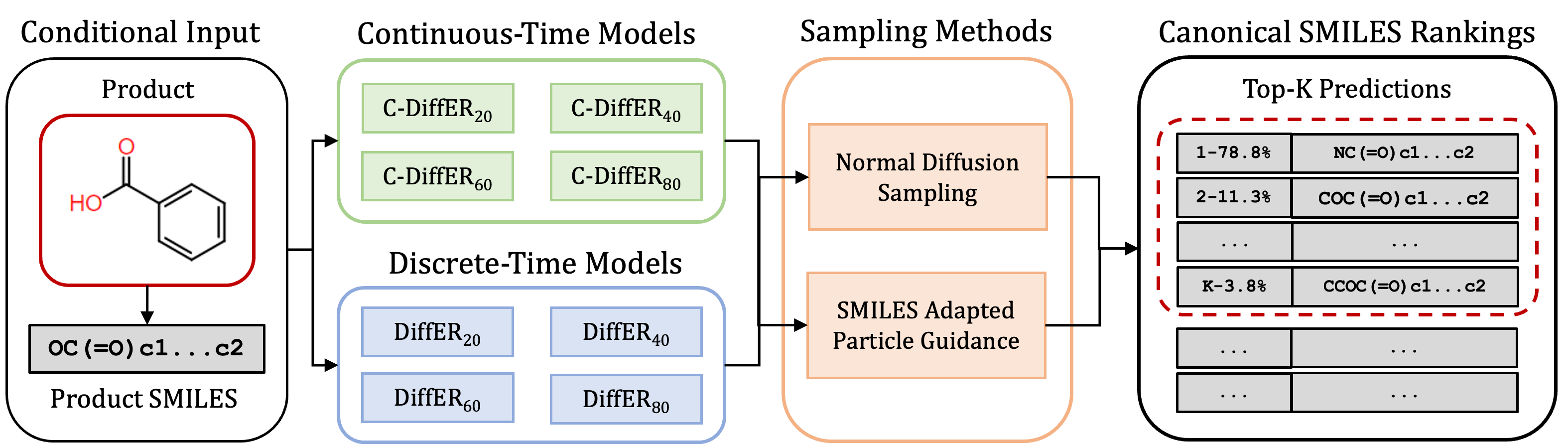}
    \caption{Overview of the DiffER$^2$ ensemble. The DiffER$^2$ ensemble contains both continuous-time (C-DiffER) and discrete-time (DiffER) models to leverage differing capabilities between the two diffusion archetypes. Both normal sampling methods and SMILES adapted particle guidance are used to generate output. Output SMILES are ranked according to frequency of the canonical form. DiffER$_\kappa$ indicates the value of $\kappa$ as described in Section~\ref{sec:differ}.}
    \label{fig:schema}
\end{figure}

In this work, we extend upon the work of DiffER \citep{current2024differ} by adapting their approach to include continuous-time discrete diffusion models and applying classifier-based particle guidance \citep{corso2023particle} to SMILES string generation. Both of these changes aim to improve the efficacy and output diversity of models over the original DiffER. A visual overview of the DiffER$^2$ ensemble approach is presented in Figure~\ref{fig:schema}. The main improvements on the original DiffER methodology are as follows:
\begin{enumerate}
    \item We incorporate continuous-time discrete diffusion models into the DiffER ensemble to leverage different capabilities of discrete-time and continuous-time diffusion.
    \item We explore the efficacy of SMILES-adapted guidance mechanisms to further enhance the diversity and accuracy of diffusion models for chemical retrosynthesis.
    \item Finally, we offer an in-depth exploration into the differing performance capabilities of auto-regressive and discrete-space diffusion models on the chemical retrosynthesis task.
\end{enumerate}
 By combining both continuous-time and discrete-time diffusion models with classifier-based particle guidance \citep{corso2023particle}, we significantly improve ensemble performance and reduce oversampling of posterior peaks while maintaining or boosting top-k accuracy across the board. Our results enhance the efficacy of diffusion ensembles for SMILES generation while providing an empirical analysis of diffusion performance compared to auto-regressive approaches for sequence generation.

\section{Related Work}

\subsection{Retrosynthesis Prediction}

Data-driven models for retrosynthesis prediction are commonly divided into three approaches: template-based, semi-template, and template-free methods. Common template-based approaches include GLN \citep{dai2019retrosynthesis} and LocalRetro \citep{chen2021deep}. These approaches primarily focus on identifying locations in a molecule where reaction templates can be applied and ranking the efficacy and feasibility of those templates on the target area. Semi-template methods include the likes of GraphRetro \citep{somnath2021learning}, G$^2$Retro \citep{chen2023g}, and GDiffRetro \citep{sun2025gdiffretro}. These approaches generally work in a two-step manner: in the first step, a reaction center is first predicted from the product molecule, and the molecule is subsequently split into synthons. In the second step, each synthon is completed to construct a final reactant molecule. Notably, GDiffRetro \citep{sun2025gdiffretro} uses discrete-time diffusion processes to complete predicted synthons. Finally, template-free methods such as Chemformer \citep{irwin2022chemformer}, MEGAN \citep{sacha2021molecule}, R-Smiles \citep{zhong2022root}, and DiffER \citep{current2024differ} directly construct reactant molecules from the product molecule. This is most often done in a sequence-to-sequence manner using SMILES strings \citep{irwin2022chemformer, zhong2022root, current2024differ}, though some approaches directly operate on graphs, such as the case of MEGAN \citep{sacha2021molecule}, which iteratively builds the reactant molecule(s) through a series of edits. DiffER \citep{current2024differ} is the first approach to use conditional discrete diffusion to accomplish the retrosynthesis prediction task, although their methodology suffers from reduced output diversity and high computational cost compared to auto-regressive decoding strategies. We primarily focus on template-free methods for performance comparison.

\subsection{Discrete-Space Diffusion}

Generative diffusion models have been a significant topic of research in recent years, and have achieved significant success in the realms of image generation \citep{croitoru2023diffusion, yang2023diffusion} and signal processing \citep{yang2023diffusion}. Though diffusion models traditionally operate in the continuous regime \citep{sohl2015deep, song2019generative, song2020score, ho2020denoising, saharia2022photorealistic, gu2022vector, zhang2023adding}, significant effort has been made extending diffusion models to discrete spaces \citep{hoogeboom2021argmax, austin2021structured, gong2022diffuseq, gong2023diffuseq, he2022diffusionbert, yuan2022seqdiffuseq, dieleman2022continuous, ghazvininejad2019mask}. Discrete-space models have demonstrated remarkable performance for molecular generation \citep{xu2022geodiff, alakhdar2024diffusion, hoogeboom2022equivariant, xu2023geometric, schneuing2024structure} and protein design \citep{gruver2023protein, watson2022broadly, ni2023generative}, but have struggled to out-compete their auto-regressive counterparts on discrete sequence generation. Furthermore, most work in this space has primarily focused on discrete-time diffusion. Only recently has discrete-space diffusion been extended to continuous-time implementations \citep{campbell2022continuous, zhao2024unified, xu2024discrete}, which demonstrate improved modeling capabilities and sampling efficiency compared to their discrete-time parallels. Furthermore, incorporating recent advances in guidance mechanisms during diffusion sampling \citep{bansal2023universal, liu2023more, epstein2023diffusion, ho2022classifier} remains a challenge, due to the difficulty of applying gradient-based approaches on a discrete domain \citep{schiff2024simple, gruver2023protein}. \citet{schiff2024simple} recently adapted classifier-free and classifier-based guidance mechanisms for discrete diffusion and demonstrated promising results on molecular generation tasks, though more work needs to be done constructing and evaluating guidance mechanisms for discrete diffusion tasks.

\section{Methods}
\label{sec:methods}

Following the original DiffER \citep{current2024differ}, we treat retrosynthesis prediction as a discrete sequence-to-sequence generation task, where the target product is a conditional input to the diffusion generation akin to text-to-image models. We employ both discrete-time and continuous-time diffusion models. Like DiffER, we incorporate length prediction strategies to improve diffusion sampling as well as enable robust model ensembling. Finally, we adapt classifier-based particle guidance \citep{corso2023particle} to SMILES sequence generation to improve output diversity and reduce over-sampling of posterior peaks. By applying our techniques to SMILES generation, we are able to offer direct comparisons to prior works, and more clearly demonstrate the effectiveness of our diffusion ensembling and guidance methodologies. However, we recognize the existence of other molecular string representations such as SELFIES \citep{selfies}. We offer a comparison to a SEFLIES-based implementation of our work in Appendix \ref{app:full_topk}.

\subsection{Preliminaries}

Following the notation of \citet{zhao2024unified}, let $x_0 \sim p_{data}(x_0)$ be a categorical random variable representing the observed data with discrete distribution $p_{data}(x_0; K)$. Let $x_t \sim q(x_t)$ be the latent variable at time $t \in [0, 1]$ (continuous-time, max time $T=1$) or $t \in \{1,...,T\}$ (discrete-time) and $x_{t|s} \sim q(x_t | x_s)$ be the conditional random variable. Let $x_t^{1:L_x} \in \{0, 1\}^{L_x \times K}$ be a sequence of length $L_x$ at time $t$, and let $x_t^{[\ell]}$ be the sequence indexed at position $\ell$. We assume each discrete random variable is governed by a categorical distribution $\mathcal{C}$ with probability distribution $p \in [0, 1]^K$, where $||p||_1 = 1$, such that $x_t \sim \mathcal{C}(x_t ; p)$. Furthermore, let $Q_{t|s} \in [0,1]^{K \times K}$ represent the transition matrix between times $s$ and $t$, such that $[Q_{t|s}]_{ij} = q(x_t = e_i| x_s = e_j)$, where $e_i$, $e_j \in \{0, 1\}^K$ are one-hot encoded vectors.

In the forward diffusion process, noise is added to observed data until it is indistinguishable from a stationary noise distribution. \citet{zhao2024unified} show that the forward diffusion process for discrete-time and continuous-time diffusion can be unified 
into a single forward diffusion process defined by the accumulated transition matrix $\bar{Q}_{t|s}$, with $t>s$:
\begin{equation}
    \bar{Q}_{t|s} = \bar{\alpha}_{t|s}I + (1 - \bar{\alpha}_{t|s})1m^\top,
\end{equation}
where $m$ is a known probability distribution vector such that $x_T \sim \mathcal{C}(x_T ; m)$, $1$ is vector of $1$s, and  $\bar{\alpha}_{t|s}$ is the accumulated noise schedule between time $t$ and time $s$.
% In the discrete case, $\bar{\alpha}_{t|s} = \prod_{i=s+1}^t \alpha_i$ for some schedule of $\alpha_t \in [0, 1]$ such that $\lim_{t\rightarrow T}\bar{\alpha}_{t|0} = 0$. In the continuous case, $\bar{\alpha}_{t|s}$ can be recovered from the the rate matrix $R_t=\beta(t)R_b$, where $R_b\in \mathbb{R}^{K\times K}$ is the base time-independent rate matrix. $R_t$ defines how the transition matrix $Q_t$ changes with time: choosing $R_b = 1m^\top - I$ and $\bar{\alpha}_{t|s}=\exp\left(-\int_s^t \beta(a)da\right)$ results in the same solution for $\bar{Q}_{t|s}$ as before \citep{zhao2024unified}. 

In the backward diffusion process, diffusion models aim to generate samples from $p_\theta(x_s|x_t)$ for some $0 < s < t \leq T$. This is commonly accomplished by taking $p_\theta(x_s|x_t) \approx q(x_s|x_t, x_0)$, where $q(x_s|x_t, x_0)$ is the $s$-step posterior distribution \citep{hoogeboom2021argmax, austin2021structured}. Because $x_0$ is unknown, it is approximated by training a neural network model $f_t^\theta$ such that $p_\theta(x_0 | x_t) = \mathcal{C}(x_0; f_t^\theta(x_t))$. Thus, $p_\theta(x_s|x_t)$ can be estimated by sampling $x_s \sim q\left(x_s|x_t, f_t^\theta(x_t)\right)$. Analytically, this can be calculated as
\begin{equation}
    q\left(x_s|x_t, f_t^\theta(x_t)\right) = \mathcal{C}\left(x_s; \frac{\bar{Q}_{t|s}x_t \odot \bar{Q}_{s}^\top f_t^\theta(x_t)}{x_t^\top \bar{Q}_{t}^\top f_t^\theta(x_t)}\right).
\end{equation}
This posterior calculation can be further reduced to simplify computation \citep{zhao2024unified}, which we follow in our work as well. We refer readers to \citet{zhao2024unified} for a more detailed overview of the unified diffusion methodology.

% Furthermore, \citet{zhao2024unified} show that the backward diffusion process is also shared between discrete-time and continuous-time diffusion models. The backward diffusion process aims to sample from $p_\theta (x_s|x_t)$ for some $0 < s < t \leq T$. Suppose $f_t^\theta$ is a neural network model trained such that $p_\theta(x_0 | x_t) = \mathcal{C}(x_0; f_t^\theta(x_t))$. Zhao et al. show that $p_\theta (x_s|x_t)$ can then be approximated by
% \begin{equation}
%     p_\theta (x_s|x_t) = \mathcal{C}\left(x_s; (1-\mu_{t|s})\cdot f^\theta_t(x_t) + (\mu_{t|s}\bar{\alpha}_{t|s} + \gamma_{t|s}^\theta)\cdot x_t + (\mu_{t|s}(1-\bar{\alpha}_{t|s} - \gamma_{t|s}^\theta)\cdot m\right),
% \end{equation}
% where $\mu_{t|s}=\frac{1-\bar{\alpha}_s}{1-\bar{\alpha}_t}$, $\gamma^\theta_{t|s}=(\mu_{t|s} - \lambda_{t|s} - \mu_{t|s}\bar{\alpha}_{t|s})\langle f_t^\theta(x_t), x_t\rangle$, and $\lambda_{t|s} = \frac{(1-\bar{\alpha}_s)(1-\bar{\alpha}_{t|s})\langle m, x_t\rangle}{\bar{\alpha}_t + (1-\bar{\alpha}_t)\langle m, x_t\rangle}$. Because this approximation holds true for all $0 < s < t \leq T$, the backward diffusion process is also shared between discrete-time and continuous-time diffusion models.

The primary point of difference between discrete-time and continuous-time diffusion models now lies in the loss functions. We train both continuous-time and discrete-time diffusion models using the simplified variational lower bound loss functions provided by \citet{zhao2024unified} alongside a cross-entropy loss.

\subsection{DiffER Adapted Unified Diffusion}
\label{sec:differ}

DiffER \citep{current2024differ} incorporates the target product as an additional constraint to the diffusion generation process, akin to text-to-image generative diffusion models. Let $y_0^{1:L_y} \in \{0, 1\}^{L_y \times K}$ be the target product sequence associated with the reactant sequence $x_0^{1:L_x}$. Here, both $y_0$ and $x_0$ are SMILES strings representing molecular structures, with $x_0$ containing one or more concatenated reactant SMILES. Because the target product is known in retrosynthesis prediction, The diffusion problem can be parameterized such that
\begin{equation}
    p_\theta(x_0 | x_t, y_0) = \mathcal{C}(x_0; f_t^\theta(x_t, y_0)).
\end{equation}
This formulation guides the diffusion process by incorporating the target product as an additional input to the model. Here, $f_t^\theta$ usually takes the form of an encoder-decoder model, where the encoder processes $y_0$ and the decoder acts as the diffusion component on $x_t$.

Notably, one issue remains in that the length of the reactant sequence $L_x$ is unknown a priori, which is necessary to properly instantiate $x_T \sim \mathcal{C}(x_T ; m)$. Inspired by MaskPredict \citep{ghazvininejad2019mask}, DiffER \citep{current2024differ} incorporates a length prediction component into the encoder half of $f_t^\theta$, such that $L_x$ can be estimated using some model $f_\ell^{\phi}: \{0, 1\}^{L \times K} \rightarrow \mathbb{R}^{L}$, so that $L_x \approx \arg \max f_\ell^\phi(y_0)$, where, $L$ is an arbitrary maximum sequence length such that $L_x < L$ and $L_y < L$ for all $x_0, y_0$ in the data distribution. For the chemical retrosynthesis task, $L_x$ and $L_y$ are often highly correlated: thus, to encourage active learning, $f_\ell^\phi$ is instead trained to predict the difference between $L_x$ and $L_y$, i.e.,
\begin{equation}
    L_x = L_y + (\arg \max f_\ell^\phi(y_0) - \Gamma),
\end{equation}
where $\Gamma < L$ is a sufficiently large constant allowing $f_\ell^\phi$ to effectively predict negative length changes. In practice, we set $\Gamma = \frac{L}{2}$, noting that the change in length will never be larger than half of the product sequence.

\citet{current2024differ} show experimentally that if $f_\ell^{\phi}$ is a perfect length-prediction model, then discrete-time diffusion outperforms all other methodologies for top-1 predictions by a significant margin, exhibiting that sequence length information can be highly informative when predicting reactant SMILES. However, the authors note that in practice, $f_\ell^{\phi}$ cannot be sufficiently trained to accomplish such a notable performance increase due to 1.) the distribution of $(L_x - L_y)$ is heavily biased, 2.) the SMILES encoding of a molecule can produce sequences of varying length, forcing $f_\ell^\phi(y_0)$ to learn a one-to-many mapping, and 3.) accurate length prediction requires complex understanding of chemical reagents, molecular reactivity, and molecular structure, and is itself a nontrivial problem. 

To remedy problems $1.$ and $2.$, DiffER \citep{current2024differ} randomly appends $L_p \sim \mathcal{U}\{1, \kappa\}$ padding tokens to the SMILES representations of reactants, where $\kappa$ is a set hyperparameter. This padding procedure smooths the distribution of $(L_x - L_y)$ and allows flexibility in the output size of generated reactants. A detailed overview of this procedure is provided in Appendix \ref{app:padding}. To reduce reliance on the choice of $\kappa$, DiffER trains multiple diffusion models $f_t^{\theta_\kappa}$ with different $\kappa$, producing an ensemble of models which generate possible reactants. This approach provides better coverage of the posterior distribution than a single model. We follow this same approach, but additionally incorporate continuous-time models and classifier-based guidance sampling into the ensemble.

\subsection{SMILES Adapted Particle Guidance for Chemical Generation}
\label{sec:guidance}

Aside from the incorporation of continuous-time diffusion models into the DiffER methodology and the formalization of said methodology, our other primary motivation of this work is to explore the usage of classifier-based guidance mechanisms on the DiffER ensemble models to improve output diversity. Classifier-based guidance steers diffusion sampling toward a specific classification $c$ by sampling from a tempered distribution $p^\gamma(x_s | c, x_t) \propto p(c | x_s)^\gamma p_\theta(x_s | x_t)$, where $p_\theta(x_s | x_t)$ is the diffusion model, $p(c | x_s)$ is a classifier, and $\gamma$ is the temperature parameter \citep{schiff2024simple}:
\begin{equation}
    \log p^\gamma(x_s | c, x_t) = \gamma \log p(c | x_s, x_t) + \log p(x_s | x_t).
\end{equation}
\citet{schiff2024simple} formalize classifier-based guidance for discrete-space diffusion specifically, with the assumption that $p^\gamma(x_s | c, x_t)$ factorizes independently across tokens. We follow their work to apply classifier-based guidance to our models.

\subsubsection{SMILES Adapted Particle Guidance}
One of the primary limitations of employing diffusion models for highly conditioned molecular generation tasks lies in the over-sampling of posterior peaks \citep{current2024differ, sun2025gdiffretro}, resulting in reduced diversity and number of unique generated molecules. To this end, we take inspiration from \citet{corso2023particle} to encourage greater output diversity during sampling through guidance. Corso et al. apply the gradient of a potential function $\log \Phi_t(x_t^{(1)},x_t^{(2)},...,x_t^{(n)})$ to encourage greater output diversity, where $x_t^{(i)}$ are independent samples from the diffusion model at time $t$. The potential function is constructed such that higher scores are given to distant point pairs:
\begin{equation}
    \label{eq:particle}
    \log \Phi_t(x_t^{(1)},x_t^{(2)},...,x_t^{(n)}) = \frac{\bar\alpha_t}{2}\sum_{i,j}\mathcal{K}(x_t^{(i)}, x_t^{(j)}),
\end{equation}
where $\mathcal{K}$ is a pairwise similarity kernel. Furthermore, the potential function is weighted by the noise schedule $\bar\alpha_t$ such that the guidance effect is stronger at earlier timesteps rather than later ones. Unfortunately, it is non-trivial to construct a differentiable similarity kernel between noised SMILES strings due to the non-singular nature of SMILES representations: the same molecule may be represented by many different SMILES sequences of various atom orderings.

To remedy this, we construct the potential function such that it maximizes the probability that $X_t = \{x_t^{(1)},x_t^{(2)},...,x_t^{(n)}\}$ are noisy variants of different molecules,. Let $m_0^{(i)}$ be the molecular form of the denoised sequence $x_0^{(i)}$. Then we aim to maximize the probability
\begin{equation}
    \bigcup_{i \neq j} p(m_0^{(i)} \neq m_0^{(j)} | x_t^{(i)}, x_t^{(j)}) = \prod_{i\neq j} p(m_0^{(i)} \neq m_0^{(j)} | x_t^{(i)}, x_t^{(j)}),
\end{equation}
Which factors as a result of the independence of generated SMILES. Taking the $\log$, we can construct the potential function as
\begin{equation}
    \log \Phi_t(X_t) = \sum_{i \neq j} \log p(m_0^{(i)} \neq m_0^{(j)} | x_t^{(i)}, x_t^{(j)}).
\end{equation}
We train a classifier to estimate $p(m_0^{(i)} \neq m_0^{(j)} | x_t^{(i)}, x_t^{(j)})$. Recognizing that the parameterized diffusion model $f_t^\theta$ is already trained to produce an estimate of the de-noised sequences, we train an additional classifier $f_\mathcal{K}^{\Phi}$ to detect if $\hat{x}_{0,t}^{(i)} = f_t^\theta(x_t^{(i)})$ and $\hat{x}_{0,t}^{(j)} = f_t^\theta(x_t^{(j)})$ represent the same denoised molecular structure, i.e.,
\begin{equation}
    p(m_0^{(i)} \neq m_0^{(j)} | x_t^{(i)}, x_t^{(j)}) = f_\mathcal{K}^\Phi (\hat{x}_{0,t}^{(i)}, \hat{x}_{0,t}^{(j)}) = (f_\mathcal{K}^\Phi \circ f_t^\theta)(x_t^{(i)}, x_t^{(j)}).
\end{equation}
This parameterization results in the final potential function
\begin{equation}    
    \log \Phi_t(X_t) = \sum_{i\neq j}\log\left((f_\mathcal{K}^\Phi \circ f_t^\theta)(x_t^{(i)}, x_t^{(j)})\right).
\end{equation}

\subsubsection{Applying Potential Functions via Classifier-based Guidance}
Rather than using this potential function directly, we instead construct a classifier on $x_t^{(i)}$ alone, recognizing that applying the classifier individually for all $x_t^{(i)} \in X_t$ reproduces the potential function:
\begin{equation}
    \label{eq:classifier}
    \log p_\Phi(\mathbb{I}[\hat m_{0,t}^{(i)} \notin \hat M_t/\{\hat m_{0,t}^{(i)}\}]| x_t^{(i)}, X_t) = \sum_{j\neq i}\log\left((f_\mathcal{K}^\Phi \circ f_t^\theta)(x_t^{(i)}, x_t^{(j)})\right).
\end{equation}
Here, $\hat m_{0,t}^{(i)}$ is the molecular form of $\hat x_{0,t}^{(i)}$, $\hat M_t$ is the set of molecules $\{\hat m_{0, t}^{(1)},...,\hat m_{0,t}^{(n)}\}$ associated with $X_t$, and $\mathbb{I}$ is the indicator function. For the sake of clarity, we will refer to $\mathbb{I}[\hat m_{0,t}^{(i)} \notin \hat M_t/\{\hat m_{0,t}^{(i)}\}]$ as the guidance class $c$. We add the further stipulation that if $\hat x_{0,t}^{(i)}$ is not a valid SMILES string or if $\hat m_{0,t}^{(i)}$ is not a valid molecule, then the indicator function defaults to False.

We finalize our guidance strategy by follow the approach of \citet{schiff2024simple} to apply classifier guidance to discrete sequences. This results in the posterior estimate
\begin{equation}
    \label{eq:posterior}
    \log p^\gamma_{\Phi,\theta}(x_s|c, x_t) = \sum_{\ell = 1}^L \log \left(\frac{p_\Phi(c|\tilde x)^\gamma p_\theta (x_s^{[\ell]}|x_t)}{\sum_{\tilde{x}}p_\Phi(c|\tilde x)^\gamma p_\theta (x_s^{[\ell]}|x_t)}\right), 
\end{equation}
where $\tilde{x} = [x_t^{1:\ell-1}, x_s^\ell, x_t^{\ell+1:L}]$ is the sequence such that $\tilde{x}$ and $x_t$ are equal at all positions $\ell^\prime \neq \ell$ and $\tilde x$ and $x_s$ are equal at position $\ell$. We set $\gamma=\bar\alpha_t$ to weight the guidance effect similarly to Equation~\ref{eq:particle}. Finally, we use the first-order Taylor approximation of $\log p_\Phi$ to efficiently compute the classifier $p_\Phi(c|\tilde x)$ using only a single forward pass \citep{schiff2024simple}:
\begin{equation}
    \label{eq:taylor}
    \log p_\Phi(c| \tilde x) = (\tilde x - x_t)^\top \nabla_{x_t}\log p_\Phi(c | x_t) + \log p_\Phi(c | x_t). 
\end{equation}
We refer readers to \citet{schiff2024simple} for detailed proofs of this result. 

Training and implementation details for $f_\mathcal{K}^\Phi$ are provided in Appendix~\ref{app:particle_training}. During inference, $f_\mathcal{K}^\Phi$ is incorporated in a manner reminiscent of reinforcement learning, allowing the model to learn and adapt throughout the diffusion process.

\section{Experiments}
\label{sec:metrics}

\paragraph{Experimental Setup.}
We train and test our models on the USPTO-50K dataset \citep{Lowe2017}, a commonly used set of patented reactions that provides SMILES strings for both reactants and products and widely regarded as the benchmark dataset for retrosynthesis prediction. Reaction solvents are not included in the retrosynthesis task. Each pair of products and reactants is augmented 20 times according to \citet{zhong2022root}, producing a final dataset of 1M root-aligned SMILES pairs. We use the augmented train, validation, and test sets provided by Zhong et al.\footnote{\url{https://github.com/otori-bird/retrosynthesis}}

The DiffER$^2$ ensemble contains eight individually trained diffusion models, depicted in Figure~\ref{fig:schema}. Each model shares the same network architecture, differentiated only by the diffusion parameters and padding augmentation described in Section~\ref{sec:differ}. Four of the diffusion models follow the original DiffER training \citep{current2024differ} with losses adjusted according to \citet{zhao2024unified}: these models operate in discrete-time with padding parameter $\kappa \in \{20, 40, 60, 80\}$. The remaining four models operate in continuous-time, each paralleling their discrete-time counterparts with a padding parameter $\kappa \in \{20, 40, 60, 80\}$. Further details on model architecture are provided in Appendix~\ref{app:archs}.
% All eight models take the form of standard encoder-decoder transformers with 6 layers, 8 attention heads, a latent size of 512, a feed-forward layer of size 2048, and GeLU activation \citep{hendrycks2016gaussian}. All models are trained using an Adam optimizer with learning rate $0.0001$ and a dropout rate of $0.1$ for 90 epochs. Both discrete-time and continuous-time models are trained using a log-linear noise schedule and are sampled over 100 diffusion timesteps. All models are trained on a single Nvidia A6000 GPU with 48GB memory.
Models which include particle guidance described in Section \ref{sec:guidance} are denoted DiffER$^2$PG and DiffER$^2$PG+. In DiffER$^2$PG, guidance sampling is run instead of the standard sampling procedure. In DiffER$^2$PG+, guidance sampling is run alongside standard sampling procedures, resulting in an output distribution that combines DiffER$^2$ and DiffER$^2$PG.
% We also report select results for two non-ensemble continuous-time diffusion models in the appendices. These models are denoted C-DiffER and C-DiffER$_{20}$. C-DiffER does not use the length prediction methodology described in section \ref{sec:differ} and is used as a baseline comparison to a non-ensemble diffusion model. C-DiffER$_{20}$ uses the length prediction methodology with $\kappa = 20$.

\paragraph{Evaluation Metrics.}
We evaluate our models using standard top-$k$ accuracy metrics. The DiffER$^2$ ensemble is run on each product in the test set and generated reactants are ranked by frequency, with the most generated reactant having the highest ranking. The top-$k$ accuracy is defined as the proportion of reactions in which the ground-truth reactant is in the top-$k$ ranks. Top-k accuracy is reported for $k \in \{1, 3, 5, 10\}$. We compare the DiffER$^2$ ensemble with a wide range of template-based, semi-template, and template-free methods for retrosynthesis prediction. However, we primarily focus on comparing performance against other template-free methods. In addition to top-k accuracy, we include results for round-trip accuracy using a forward synthesis model in Appendix~\ref{app:rt_accuracy} and provide a case study on a ring-forming reaction in the main results and three other reactions in Appendix~\ref{app:case_study}.

\paragraph{Property-Specific Performance Analysis.} We additionally analyze how model performance compares against various structural properties of the SMILES sequences to better understand model performance. These properties include the target length difference between product and reactant SMILES, the edit distance between product and reactant SMILES, if the target reaction is ring adding/ring removing, and the number of branches in the target SMILES, each detailed in ~Appendix\ref{app:property_coeffs}. For each property, we calculate the slope and significance of a regression between the property statistic and the top-$k$ accuracy. In addition to these results, we report the performance of DiffER$^2$PG across different reaction types in Appendix~\ref{app:reaction_types}. We analyze how different molecular structures correlate with model performance in Appendix~\ref{app:fingerprints}. Finally, we analyze the attention maps for a non-ensemble diffusion model in Appendix~\ref{app:attention}.
 
\section{Results}

\begin{figure}[b]
    \centering
    \includegraphics[width=0.8\linewidth]{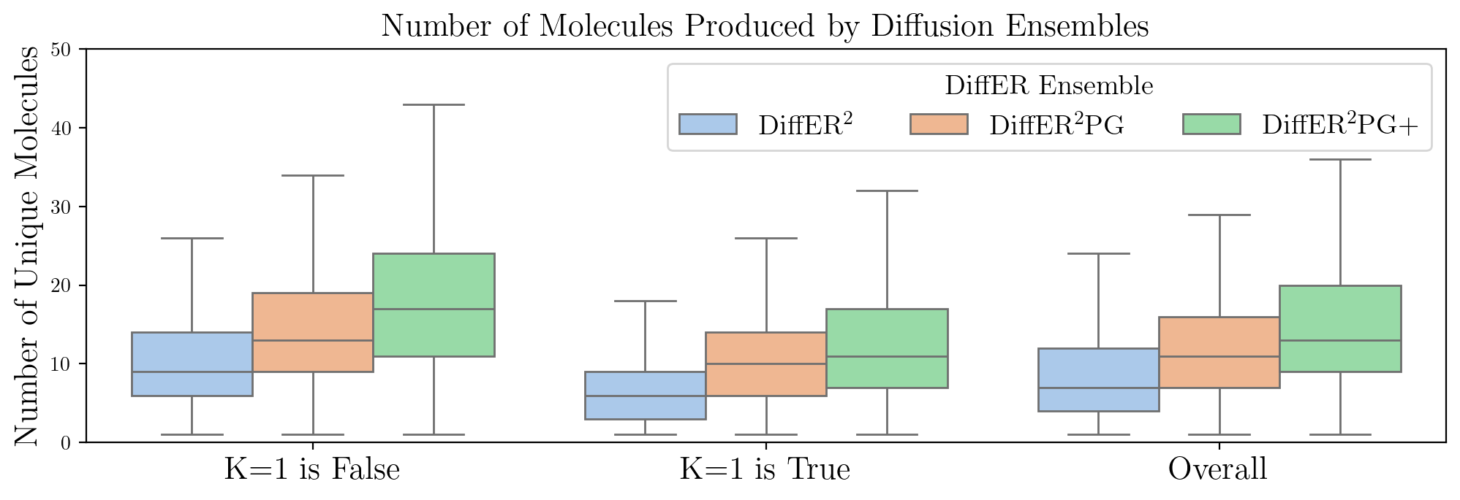}
    \caption{Box plot showing the number of unique molecules produced by the various DiffER$^2$ ensembles. "K=1 is False" indicates reactions that do not have an accurate top-1 reactant. "K=1 is True" indicates reactions that do have an accurate top-1 reactant. "Overall" includes all reactions.}
    \label{fig:num_samples}
\end{figure}

Top-K accuracies for the DiffER$^2$ ensemble are reported in Table~\ref{tab:topk}. DiffER$^2$PG+ achieves the best top-1, top-3, and top-5 (statistically tied with R-SMILES) among template-free models and the second-best top-10 accuracy behind R-SMILES \citep{zhong2022root}. R-SMILES achieves the best top-5 and top-10 performance, but lags behind DiffER$^2$PG+ for $ k=1, 3$. DiffER$^2$PG+ may better estimate posterior peaks for SMILES sequences than R-SMILES, but it still struggles to sample from lower probability regions in the posterior. This result is paralleled in GDiffRetro \citep{sun2025gdiffretro}, a semi-template method that utilizes diffusion sampling techniques. GDiffRetro achieves remarkable $k=1$ accuracy, the highest of all baseline models, but struggles to produce strong results for higher values of $k$ compared to other methods. \citet{sun2025gdiffretro} suggest this may be due to the oversampling of posterior peaks, which is also observed in our models. By employing an ensemble of diffusion models and incorporating SMILES-adapted particle guidance, DiffER$^2$PG+ can mitigate some of this bias and provide greater coverage of the posterior, as evidenced by our stronger performance for $k>1$ compared to GDiffRetro.
Compared to the original DiffER~\citep{current2024differ}, the DiffER$^2$ ensembles all achieve stronger results than the discrete-time only ensemble, particularly for higher values of $k$. This suggests that including continuous-time models greatly improves the diversity of output reactants. Notably, there is minimal difference between all DiffER methodologies for top-1 accuracy, indicating that the location of posterior peaks for the various ensembles may be similar, but that the shape of the distribution for lower probability samples is different. Interestingly, the top-k performances of DiffER$^2$ and DiffER$^2$PG are remarkably similar despite the latter ensemble producing more reactants.

Figure~\ref{fig:num_samples} shows box plots of the unique reactants generated by each ensemble for products in the test set. The greater number of unique molecules produced by DiffER$^2$PG indicates that the SMILES-adapted particle guidance methodology does result in greater output diversity and reduced sampling of posterior peaks. The combined result of DiffER$^2$PG+ has an even greater number of unique molecules, indicating that the difference between sample sets generated from DiffER$^2$ and DiffER$^2$PG is substantial. Notably, this result may be impacted by the limited number of samples taken from each model in the ensemble, which may not reflect the full posterior distribution for each model. Regardless, DiffER$^2$PG is an effective approach to improving output diversity while maintaining comparable performance to DiffER$^2$. This is desirable because multiple reactants could produce the same product, thus validating accuracy, and output diversity is critical. Like \citet{current2024differ}, we observe that fewer molecules are sampled when the top-1 result is accurate, indicating that the models still over-sample from posterior peaks to the deficit of output diversity. However, these effects are lessened compared to DiffER.

\begin{table}[t]
    \centering
    \caption{Top-K accuracy for a select set of baseline models. A full table of results can be found in Appendix~\ref{app:full_topk}. The best performing model in each category is \textbf{bolded}, while the second-best performing model is \underline{underlined}. A $\dagger$ indicates results that are statistically significant from DiffER at the 95\% level.}
    \begin{tabular}{l l l l l l}
        \toprule
         Category & Model & K=1 & 3 & 5 & 10 \\
        \midrule
        \multirow{2}{*}{Template-based} & GLN \citep{dai2019retrosynthesis} & \underline{52.5} & \underline{69.0} & \underline{75.6} & \underline{83.7} \\
                                           & LocalRetro \citep{chen2021deep} & \textbf{53.4} & \textbf{77.5} & \textbf{85.9} & \textbf{92.4} \\
        \midrule
        \multirow{3}{*}{Semi-template} & RetroPrime \citep{wang2021retroprime} & 51.4 & 70.8 & 74.0 & \underline{76.1} \\
                                          & G$^2$Retro \citep{chen2023g} & \underline{53.9} & \underline{74.6} & \underline{80.7} & \textbf{86.6} \\
                                          & GDiffRetro \citep{sun2025gdiffretro} & \textbf{58.9} & \textbf{79.1} & \textbf{81.9} & - \\
        \midrule
        \multirow{6}{*}{Template-free} & MEGAN \citep{sacha2021molecule} & 48.1 & 70.7 & 78.4 & 86.1\\
                                          % & Chemformer \citep{irwin2022chemformer} & 54.3 & - & 62.3 & 63.0 \\
                                          % & Retroformer \citep{wan2022retroformer} & 53.2 & 71.1 & 76.6 & 82.1 \\
                                          & R-SMILES \citep{zhong2022root} & 56.3 & 79.2 & \textbf{86.2} & \textbf{91.0} \\
                                          & DiffER \citep{current2024differ} & \underline{57.6} & 79.0 & 84.1 & 87.4 \\
        \cmidrule(lr){2-6}                                   
                                          & DiffER$^2$ &
                                          \underline{57.6} &
                                          \underline{79.5} &
                                          85.3$^\dagger$ &
                                          88.2 \\
                                          & DiffER$^2$PG &
                                          57.4 &
                                          79.4 &
                                          85.4$^\dagger$ &
                                          88.6$^\dagger$ \\
                                          & DiffER$^2$PG+ &
                                          \textbf{57.7} &
                                          \textbf{79.8} &
                                          \underline{86.1}$^\dagger$ &
                                          \underline{89.9}$^\dagger$ \\
        \midrule                             
    \end{tabular}
    \label{tab:topk}
\end{table}

\paragraph{Case Study.}
A case study on an N-heterocyclization reaction with dihalides for the DiffER$^2$PG+ ensemble is presented in Figure~\ref{fig:main_case}. Additional case studies are presented in Appendix~\ref{app:case_study}. The target reaction is a  reaction of  a primary amine with a dihalide in an alkaline aqueous medium to form a heterocyclic ring \citep{ju2006aqueous}. While the ensemble model predicts this reaction, it is only predicted 4\% of the time, making it the third most frequently predicted reaction. The top-ranked predicted reaction is a Borch reductive amination \citep{borch2003reductive} in which a carbonyl group is reacted with a secondary amine in the presence of a reducing agent to produce a tertiary amine. Unfortunately, due to the presence of multiple carbonyl groups in the predicted reactant, this reaction will likely produce unwanted byproducts. The second-ranked reaction is a substitution reaction with a mesylate ester and a secondary amide nucleophile. Again, this reaction does produce the desired product, but may produce byproducts due to the electrophilicity of the carbamate group, which may react with the amine nucleophile.

\begin{figure}[t]
    \centering
    \includegraphics[width=\linewidth]{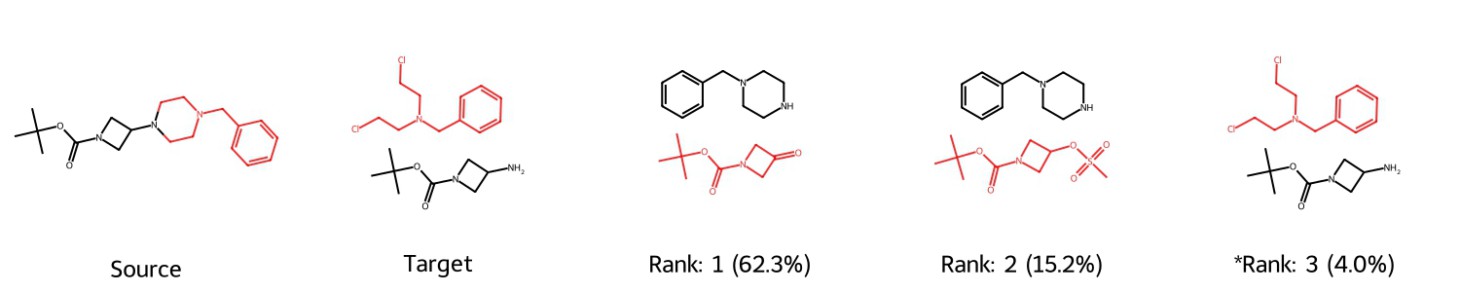}
    \caption{Case study on a N-heterocyclization reaction with dihalides. From left to right: the source product, the ground-truth reactant(s), and the top-3 predicted reactant(s) for the DiffER$^2$PG+ ensemble. Red highlighting indicates differences between the source product and the reactants. The value in parentheses indicates the rate at which the reactant(s) are produced by the ensemble.}
    \label{fig:main_case}
\end{figure}

To recap, all three top-ranked reactions could be used to form the target product, but only the 3rd reaction, which matches the ground truth, would limit the creation of byproducts. Notably, this reactant is not produced by the original DiffER ensemble \citep{current2024differ}, demonstrating that DiffER$^2$PG+ significantly improves output diversity to benefit performance.
In this case study, as well as those presented in Appendix~\ref{app:case_study}, DiffER$^2$PG+ demonstrates significant capability to predict reactants that could be used to produce the desired product, though some reactions may be suboptimal due to the possibility of undesirable side products. Thus, while the models can learn appropriate reaction mechanisms and patterns, the predicted forward reaction's purity may not be considered by the models. Future work should consider approaches that include reducing the possibility of side products as an objective. Furthermore, ambiguity in the optimality of reactants could be mitigated by providing additional guidance to the model in a human-in-the-loop setting.

\begin{figure}[h!]
    \centering
    \includegraphics[width=0.9\linewidth]{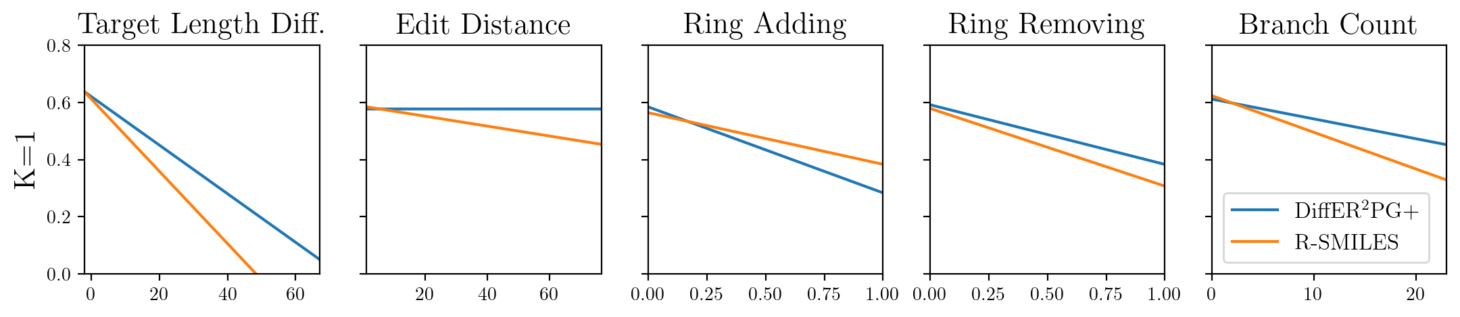}
    \caption{Lines of best fit between top-k accuracy and reactant properties for the DiffER$^2$PG+ and R-SMILES \citep{zhong2022root} models. All slopes are significantly different from 0 ($p < 0.01$) except Edit Distance.}
    \label{fig:lines_k1}
\end{figure}

\paragraph{Property Results.}
Figure~\ref{fig:lines_k1} shows the results of the property-specific performance analysis described in Section~\ref{sec:metrics} for top-1 accuracy, with additional results available in Appendix~\ref{app:property_coeffs}. These results highlight substantial differences in the capabilities of DiffER$^2$PG+ and R-SMILES. DiffER$^2$PG+ is significantly more capable at predicting reactant SMILES with a larger increase in sequence length than the auto-regressive model. This result also helps explain the increased capability of diffusion models to predict ring-removing reactions, as these are often reactions with a high increase in sequence length. This difference in capability can be understood via the attention maps for the diffusion models: rather than building up the sequence token-by-token like an auto-regressive model, the diffusion models work backward, attending to the full-sequence length in early layers then decreasing the attended length as the model goes deeper (see Appendix~\ref{app:attention}). Thus, diffusion models may be "less inclined" to produce shorter sequences.

In contrast, diffusion models seem less capable of modeling ring-adding reactions than auto-regressive models, though this difference is only significant for $k>1$. 
Both models have a negative association with the edit distance between product and reactant SMILES, except for top-1 accuracy, in which DiffER$^2$PG+ has no significant relationship with the edit distance. This suggests that some reactions with large edits remain "easy" for DiffER$^2$PG+ to predict. Both models struggle to generate sequences with highly non-sequential behavior, as evidenced by the negative relationship between performance and branch count. Overall, DiffER$^2$PG+ outperforms R-SMILES conditioned on all properties except for ring-adding reactions.

% \subsection{Limitations}

% While the DiffER$^2$ ensembles prove to be competitive models in comparison to other modeling schemes for chemical retrosynthesis, one major disadvantage of our approach lies in the computational cost of repeatedly sampling from multiple separate diffusion models. R-SMILES is able to generate top-10 results with beam search for 20 different augmentations of the test set in $\sim30$ minutes on a single Nvidia A6000 GPU. Meanwhile, running \textit{one} of the models in the DiffER$^2$ ensemble on the same augmented test data for 100 diffusion timesteps takes $\sim2$ hours for normal sampling, or $\sim6$ hours with SMILES adapted particle guidance on the same Nvidia A6000 GPU. Notably, however, retrosynthesis prediction is not often run on a large number of target products at once, but rather on a small set of target products. If we divide the generation time of the normal sampling by the size of the test set, a model in the DiffER$^2$ ensemble takes $\sim2$ seconds to generate samples for a single product with $20\times$ augmentation. While this is a sizable increase in computation time over the vanilla R-SMILES, it remains fast enough to be used in practice.

\section{Conclusion}

In this work, we introduce SMILES-adapted particle guidance to encourage output diversity in diffusion models for SMILES generation on the chemical retrosynthesis task. Combining this guidance mechanism with the ensemble approach of DiffER \citep{current2024differ}, we can significantly increase the number of unique molecules generated while improving top-k accuracy metrics on benchmark datasets. Furthermore, we offer an in-depth empirical analysis on the capabilities and limitations of discrete diffusion for SMILES generation compared to traditional auto-regressive approaches. In future work, we aim to investigate additional guidance methods for SMILES diffusion models to reduce byproducts and explore applications of rectified linear flows for SMILES generation.

\subsubsection*{Reproducibility}
We detail all model architectures and hyperparameters used to produce DiffER$^2$ in Appendix \ref{app:archs}. We follow the methods outlined in Section \ref{sec:methods} and Appendix \ref{app:particle_training} alongside the experimental setup in Section \ref{sec:metrics} to train and evaluate our models. All code used to produce the results of DiffER$^2$ will be made publicly available.


% This project was made possible, in part, by support from the National Science Foundation grant no. IIS-2133650 and the National Library of Medicine grant no. 1R01LM014385. Any opinions, findings and conclusions or recommendations expressed in this paper are those of the authors and do not necessarily reflect the views of the funding agency.

\bibliography{references}
\bibliographystyle{iclr2026_conference}

\newpage
\appendix
\section{Appendix}
\input{supplementary_material.tex}

\end{document}

%% file: math_commands.tex
\usepackage{amsmath,amsfonts,bm}

\def\eqref#1{equation~\ref{#1}}
\def\1{\bm{1}}

\DeclareMathAlphabet{\mathsfit}{\encodingdefault}{\sfdefault}{m}{sl}
\SetMathAlphabet{\mathsfit}{bold}{\encodingdefault}{\sfdefault}{bx}{n}

%% file: supplementary_material.tex
\subsection{Details on DiffER's Padding Procedure}
\label{app:padding}

During training, DiffER \citep{current2024differ} randomly appends padding tokens to the SMILES representations of reactants. The target reactant sequence $x_0^{1:L_x}$ is randomly padded on the right by a sequence of padding tokens with random length, such that
\begin{equation}
    x_{0}^\prime = \left[ x_{0}^{1:L_x}, e_p^{1:L_p} \right] ~\text{and}~ x_t \sim q(x_t| x_{0}^\prime),
\end{equation}
where $e_p$ is the one-hot encoder vector of the pad token and $L_p \sim \mathcal{U}\{1, \kappa\}$, with $\mathcal{U}\{1, \kappa\}$ being the discrete uniform distribution between $1$ and some constant $\kappa>1 \in \mathbb{N}$. The choice of $\kappa$ allows the diffusion model to strike a balance between effective length prediction with $f_\ell^\phi$ and allowing $f_t^\theta$ to determine the sequence length by predicting padding tokens during sampling. 

We can show that the variant padding methodology is in effect a moving average kernel on the distribution of $(L_x - L_y)$ with width $\kappa$. Let $L_{x-y}\in [-L/2, L/2]$ denote the difference $(L_x - L_y)$ with probability distribution $L_{x-y} \sim \mathcal{C}(L_{x-y}; p_{\ell})$, where the probability vector $p_\ell$ is calculated from the training data distribution. Let $L_{x-y}^\prime \sim \mathcal{C}(L_{x-y}^\prime; p_{\ell}^\prime)$ represent the difference distribution after random padding with probability vector $p_\ell^\prime$, and let $P_{i|j} = P(L_{x-y}^\prime = i|L_{x-y}=j)$ be the probability that a sequence pair with difference $j$ maps to a difference of $i$ for the padded sequence. Then
\begin{equation}
    {p_\ell^\prime}_i = \sum_{j=1}^L P_{i|j}\cdot {p_\ell}_j = \frac{1}{\kappa}\sum_{j=1}^L {p_\ell}_j \mathbb{I}[i-\kappa \leq j < i] = \frac{1}{\kappa}\sum_{j=i-\kappa}^{i-1} {p_\ell}_j.
\end{equation}
Thus, $p_\ell^\prime$ is a moving average of the previous $\kappa$ indices of $p_\ell$. In effect, the padding methodology reduces the heavy bias observed in the distribution of $L_{x-y}$ by smoothing the probability vector $p_\ell$.

\subsection{Training Details for Particle Guidance Model}
\label{app:particle_training}

\begin{algorithm}[h]
\caption{Pseudocode for particle guidance sampling procedure.}
\label{alg:generation}
\textbf{Inputs}: target product $y_0$, timesteps $T$, number of samples $N$, diffusion model encoder $f_{encoder}^\theta$, diffusion model decoder $f_{decoder}^\theta$, noise schedule $\bar\alpha_t$ \\
\textbf{Output}: Sampled reactants $\{\hat x_0^{(1)}, \hat x_0^{(2)}, ..., \hat x_0^{(N)}\}$
\begin{algorithmic}[1] %[1] enables line numbers
\FOR{$i$ in $1$ to $N$}
    \STATE Initialize $y_0^{(i)}=$ Augment$(y_0)$ according to Zhong et al.\cite{zhong2022root}
    \STATE Initialize $x_T^{(i)} = \mathcal{C}(x_T; m)$ from noise distribution
    \STATE Compute encoder memory $H_i = f_{encoder}^\theta(y_0^{(i)})$
\ENDFOR
\FOR{$t$ in $T$ to $1$}
    \STATE $s = t - 1$
    \STATE Let $X_t = \{x_t^{(1)}, x_t^{(2)},...,x_t^{(n)}\}$
    \FOR {$i$ in $1$ to $N$}
        \STATE Predict $\hat x_0^{(i)} \leftarrow f_{decoder}^\theta(x_t^{(i)}, H_i, t)$
        \STATE Compute $p_\Phi(c=1|x_t^{(i)}, X_t)$ according to equation \ref{eq:classifier}
        \STATE Compute $x_s^{(i)}$ according to equations \ref{eq:posterior} and \ref{eq:taylor}
        \STATE Update $f_\mathcal{K}^\Phi$ and $f_{encoder}^\theta$ using classifier loss (Binary Cross Entropy)
    \ENDFOR
\ENDFOR
\STATE $\hat x_0^{(i)} \leftarrow f_{decoder}^\theta(x_t^{(i)}, H_i, t)$
\STATE Return $\{\hat x_0^{(1)}, \hat x_0^{(2)}, ..., \hat x_0^{(N)}\}$
\end{algorithmic}
\end{algorithm}

The guidance model $f_\mathcal{K}^\Phi$ is trained in a reinforcement learning paradigm during inference with no explicit pretraining. Recognizing that the encoder portion of the diffusion model $f_{encoder}^\theta$ is time-independent and already adept at recognizing molecular structure, we construct $f_\mathcal{K}^\Phi$ as an additional classification head following an $f_{encoder}^\theta$. Let $H^{1:L_i}_i \in \mathbb{R}^{L_{i} \times d}$ be the hidden representation of a SMILES sequence $x^{(i)} \in \{0, 1\}^{L_i \times K}$ with hidden size $d$ and length $L_i$ obtained from the encoder model, i.e., $H_i^{1:L_i} = f_{encoder}^\theta(x^{(i)})$. Let $H^{(i)} = \frac{1}{L_i}\sum_{\ell=1}^{L_i} H_i^{[\ell]}$ be the hidden representation average over the sequence length. The classifier model is constructed as a function of the difference between the mean hidden representations for strings $x^{(i)}, x^{(j)}$ along the sequence length:
\begin{equation}
    f_\mathcal{K}^\Phi(x^{(i)}, x^{(j)}) = \sigma\left( \Phi^\top \left(H^{(i)} - H^{(j)}\right) \right),
\end{equation}
where $\Phi\in\mathbb{R}^d$ are the model parameters. By averaging over the sequence length $L_i$ to construct $H^{(i)}$, we help remove the effect of sequence ordering on the classification. This is desirable due to non-singular nature of SMILES representations.

During inference, we train $f_\mathcal{K}^\Phi$ in a manner akin to reinforcement learning. Note that $f_\mathcal{K}^\Phi$ is not explicitly pretrained aside from the encoder model $f_{encoder}^\theta$. We allow both the classifier weights $\Phi$ and the encoder weights $\theta_{encoder}$ to be optimized during sampling. Note that because the encoder model is only run once at the beginning of sampling, changing the weights of $f_{encoder}^\theta$ during future sampling steps does not directly affect the overall diffusion model $f_t^\theta$. Pseudocode for the joint training/sampling procedure for $f_\mathcal{K}^\Phi$ is provided in algorithm \ref{alg:generation}.

\subsection{Details on Model Architectures and Diffusion Parameters}
\label{app:archs}

All models in the DiffER$^2$ ensembles take the form of standard encoder-decoder transformers with 6 layers, 8 attention heads, a latent size of 512, a feed-forward layer of size 2048, and GeLU activation \cite{hendrycks2016gaussian}. All models are trained using an Adam optimizer with learning rate $0.0001$ and a dropout rate of $0.1$ for 90 epochs. Both discrete-time and continuous-time models are trained using a log-linear noise schedule and are sampled over 100 diffusion timesteps. All models are trained on a single Nvidia A6000 GPU with 48GB memory.

\subsection{Complete Table of Top-K Accuracy}
\label{app:full_topk}

\begin{table}[h!]
    \small
    \centering
    \caption{Complete top-K accuracy comparison to baseline models. DiffER$^2$ is the ensemble model and C-DiffER are the individual continuous-time diffusion models. The best performing model in each category is \textbf{bolded}, while the second-best performing model is \underline{underlined}.}
    \begin{tabular}{l l r r r r}
        \toprule
         Category & Model & K=1 & 3 & 5 & 10 \\
        \midrule
        \multirow{4}{*}{Template-based} & Retrosim \cite{coley2017computer} & 37.3 & 54.7 & 63.3 & 74.1 \\
                                           & Neuralsym \cite{segler2017neural} & 44.4 & 65.3 & 72.4 & 78.9 \\
                                           & GLN \cite{dai2019retrosynthesis} & \underline{52.5} & \underline{69.0} & \underline{75.6} & \underline{83.7} \\
                                           & LocalRetro \cite{chen2021deep} & \textbf{53.4} & \textbf{77.5} & \textbf{85.9} & \textbf{92.4} \\
        \midrule
        \multirow{5}{*}{Semi-template} & G2Gs \cite{shi2020graph} & 48.9 & 67.6 & 72.5 & 75.5 \\
                                          & GraphRetro \cite{somnath2021learning} & 53.7 & 68.3 & 72.2 & 75.5 \\
                                          & RetroXpert \cite{yan2020retroxpert} & 50.4 & 61.1 & 62.3 & 63.4 \\
                                          & RetroPrime \cite{wang2021retroprime} & 51.4 & 70.8 & 74.0 & \underline{76.1} \\
                                          & G$^2$Retro \cite{chen2023g} & \underline{53.9} & \underline{74.6} & \underline{80.7} & \textbf{86.6} \\
                                          & GDiffRetro \cite{sun2025gdiffretro} & \textbf{58.9} & \textbf{79.1} & \textbf{81.9} & - \\
        \midrule
        \multirow{18}{*}{Template-free} & Seq2Seq \cite{liu2017retrosynthetic} & 37.4 & 52.4 & 57.0 & 61.7 \\
                                          & Levenshtein \cite{sumner2020levenshtein} & 41.5 & 48.1 & 50.0 & 51.4 \\
                                          & GTA \cite{seo2021gta} & 51.1 & 67.6 & 74.8 & 81.6 \\
                                          & Graph2SMILES \cite{tu2022permutation} & 51.2 & 66.3 & 70.4 & 73.9 \\
                                          & Dual-TF \cite{sun2021towards} & 53.3 & 69.7 & 73.0 & 75.0 \\
                                          & MEGAN \cite{sacha2021molecule} & 48.1 & 70.7 & 78.4 & 86.1\\
                                          & Chemformer \cite{irwin2022chemformer} & 54.3 & - & 62.3 & 63.0 \\
                                          & Retroformer \cite{wan2022retroformer} & 53.2 & 71.1 & 76.6 & 82.1 \\
                                          & Tied transformer \cite{kim2021valid} & 47.1 & 67.2 & 73.5 & 78.5 \\
                                          & R-SMILES \cite{zhong2022root} & 56.3 & 79.2 & \textbf{86.2} & \textbf{91.0} \\
                                          & DiffER \cite{current2024differ} & \underline{57.6} & 79.0 & 84.1 & 87.4 \\
        \cmidrule(lr){2-6}                                   
                                          & C-DiffER &
                                          55.9 &
                                          73.3 &
                                          75.8 &
                                          76.8 \\
                                          & C-DiffER$_{20}$ &
                                          54.5 &
                                          72.6 &
                                          75.6 &
                                          76.5 \\
                                          & C-DiffER$_{\text{Selfies}}$ &
                                          54.3 &
                                          72.3 &
                                          75.2 &
                                          77.5 \\
                                          & DiffER$^2$ &
                                          \underline{57.6} &
                                          \underline{79.5} &
                                          85.3 &
                                          88.2 \\
                                          & DiffER$^2$PG &
                                          57.4 &
                                          79.4 &
                                          85.4 &
                                          88.6 \\
                                          & DiffER$^2$PG+ &
                                          \textbf{57.7} &
                                          \textbf{79.8} &
                                          \underline{86.1} &
                                          \underline{89.9} \\
        \midrule                             
    \end{tabular}
    \label{tab:topk_full}
\end{table}

A full table of top-k metrics is provided in table \ref{tab:topk_full}. This table includes additional baseline approaches for computerized retrosynthesis prediction from prior literature as well as results for the singular C-DiffER and C-DiffER$_{20}$ diffusion models. These results highlight the advantages of the ensemble approaches, which  produce sample distributions with higher top-k accuracy than the singular diffusion models. Notably, C-DiffER does achieve stronger performance than C-DiffER$_{20}$ (which includes the DiffER adapted length prediction described in section \ref{sec:differ}), demonstrating the inherent bias in the length prediction model, but still shows reduced performance compared to R-SMILES, indicating that a single diffusion model may be an insufficient approximation of the reactant posterior. These results may also change if a greater number of samples were taken from the C-DiffER models.

C-DiffER$_\text{Selfies}$ indicates the C-DiffER model trained using SELFIES molecular representations \citep{selfies}, an alternative to SMILES string representations. Unlike SMILES, SELFIES is 100\% robust, with every SELFIES string representing a valid molecular structure. SELFIES has been proposed as an alternative to SMILES for molecular generation tasks due to the guaranteed validity of outputs, which do not suffer from the harsh syntactic constraints of SMILES. Notably, however, the SELFIES-based implementation of C-DiffER does not seem to significantly improve performance compared to SMILES-based models, and in fact exhibits a lower top-1 accuracy than the SMILES-based version. As $k$ increases, the performances become more similar, with the SELFIES-based implementation exhibiting a slightly stronger top-10 accuracy. This indicates that the choice of string-encoding has little impact on the performance of the retrosynthetic diffusion model. This is likely due to the already high per-sample validity (89.4\%) of SMILES strings produced by C-DiffER, which limits the impact of SELFIES' guaranteed validity on the model's performance. Furthermore, these results support the findings of \citet{leon2024comparing}, which found that SMILES-based models often perform better than their SELFIES counterparts on a variety of molecular prediction tasks.

\newpage
\subsection{Round-Trip Accuracy}
\label{app:rt_accuracy}

\begin{table}[b!]
    \small
    \centering
    \caption{Top-K round-trip accuracy for select models.A $\dagger$ indicates results that are significantly different from DiffER at the 95\% level.}
    \begin{tabular}{l l l l l}
        \toprule
         Model & K=1 & 3 & 5 & 10 \\
        \midrule
        R-SMILES \cite{zhong2022root} & 53.8 & 66.7 & 73.7 & 82.1 \\
        DiffER \cite{current2024differ} & 53.8 & 64.7 & 70.1 & 75.6 \\
        \midrule
        DiffER$^2$ & 54.1 & 65.2 & 70.6 & 75.5 \\
        DiffER$^2$PG & 54.4 & 66.0 & 71.8 & 77.6 \\
        DiffER$^2$PG+ & 54.4 & 66.1 & 72.0$^\dagger$ & 78.4$^\dagger$ \\
        \midrule
    \end{tabular}
    \label{tab:rt_accuracy}
\end{table}

to further evaluate our models, we compute the top-k round-trip accuracy score using the forward model of R-Smiles \citep{zhong2022root}. The round-trip accuracy is calculated by running a forward synthesis prediction model on the predicted reactants of the retrosynthesis model and computing the rate at which the original input product is recovered. We use the forward prediction model of R-Smiles, which achieves a top-1 accuracy of 91\% on the USPTO-50K dataset. However, we note that the accuracy of the forward model is likely lower for retrosynthesis predictions due to the lack of reagent predictions in the retrosynthesis benchmark, which are generally included as input to the forward model. Due to the lack of reagent information amongst retrosynthesis models, the forward model must make assumptions about reagents used, possibly reducing the performance of the model when applied to the outputs of present retrosynthesis models.

The top-k round-trip accuracy results are presented in table \ref{tab:rt_accuracy}. These scores support the conclusions obtained from the ground-truth top-k accuracy presented in table \ref{tab:topk}: DiffER$^2$PG+ consistently obtains the highest accuracy scores among the proposed models, significantly improving upon DiffER for $k=5$ and $k=10$. Furthermore, DiffER$^2$PG+ maintains the best top-1 accuracy among the tested models, though the round-trip accuracies for $k=1$ differ only slightly. Notably, the difference in performance between DiffER$^2$ and DiffER$^2$PG is significantly greater when analyzing round-trip accuracy versus ground-truth accuracy, indicating that the output distribution obtained using SMILES-adapted particle guidance sampling maintains not only improves output diversity but also predicts quality reactions. These results further support the conclusion that the proposed DiffER$^2$ changes significantly improve the output diversity and efficacy of diffusion models for template-free retrosynthesis prediction.

\subsection{Reaction Type Performance}
\label{app:reaction_types}

\begin{table}[h]
    \centering
    \caption{Top-K accuracy metrics by reaction type for the DiffER$^2$PG+ ensemble. Support indicates the number of reactions of the associated type in the test set.}
    \begin{tabular}{l r r r r r}
        \toprule
        Reaction Type & K=1 & 3 & 5 & 10 & Support \\
        \midrule
        Heteroatom alkylation and arylation & 58.5 & 81.1 & 87.5 & 91.2 & 1516 \\
        Acyclation and related processes & 70.1 & 89.9 & 93.5 & 95.5 & 1190 \\
        Deprotections & 55.2 & 80.8 & 87.1 & 90.5 & 822 \\
        C-C bond formation & 41.3 & 64.9 & 73.9 & 81.5 & 567 \\
        Reductions & 56.6 & 78.4 & 85.5 & 89.4 & 463 \\
        Functional group interconversion & 39.6 & 56.0 & 70.3 & 75.3 & 182 \\
        Heterocycle formation & 32.2 & 52.2 & 63.3 & 66.7 & 90 \\
        Oxidations & 60.2 & 81.9 & 86.7 & 91.6 & 83 \\
        Protections & 70.1 & 89.6 & 94.0 & 95.5 & 67 \\
        Functional group addition & 78.3 & 87.0 & 87.0 & 91.3 & 23 \\
        \bottomrule
    \end{tabular}
    \label{tab:reaction_types}
\end{table}

Table \ref{tab:reaction_types} reports the results of the DiffER$^2$PG+ ensemble across reaction types as determined by Schneider et al. \cite{schneider2016s}. Notably, the performance of the ensemble model is inconsistent across reaction types. This has been similarly observed in other literature \cite{chen2023g, current2024differ}, and is an unsurprising result due to the differing degree of structural changes across different reaction types. The performance of the ensemble model for is lowest for heterocycle formation, functional group interconversion, and C-C bond formation. While all three reaction types exhibit low top-1 accuracy ($<40\%$), functional group interconversion and C-C bond formation are able to achieve moderately strong results ($>75\%$) for top-10 accuracy, a significant increase in performance. This suggests that the models are capable of learning reaction patterns associated with these reaction types, but may be unable to discern when they are appropriate to be used without additional guidance. In contrast, heterocycle formation still achieves limited performance ($\sim65\%$ accuracy) for $k=10$, indicating that the reaction patterns of heterocycle formation may be less understood by the diffusion models.

In all three cases, much of the difficulty in predicting patented reactions may be attributed to ambiguity in the goals of the single-step retrosynthesis task. Unlike multi-step retrosynthetic planning, which aims to produce a reaction pathway from a core set of available reactions, single-step retrosynthesis models are generally unaware of the availability of reactants. Furthermore, as exemplified in the case studies later presented in appendix \ref{app:case_study}, there are multiple possible reactants that could be used to produce the target product, and ambiguity in why one reactant may be preferential over another makes training a capable retrosynthesis model a daunting task. If we can develop methods to better inform the retrosynthesis model when one reaction may be beneficial over another, we may be better able to construct robust and useful retrosynthesis predictors.

In contrast to the poor performance of heterocycle formation, functional group interconversion, and C-C bond formation, all other reactions types achieve strong ($\sim 90\%$ or greater) top-10 accuracy, with some even achieving considerable top-1 accuracy ($>70\%$). Many of these reaction types utilize common or routine reactions: functional group addition, for instance, has extremely low diversity in reactants. While the target product is different for each sample, the vast majority of functional group additions in the test set are bromination reactions featuring N-bromosuccinimide as a reactant, a commonly used reagent for its bromine-donating properties. This commonality of reactants is also observed in the protections subset. In contrast, the high-support acyclation and related processes and heteroatom alkylation and arylation subsets exhibits a wider diversity of reactants, but all with a shared mechanism which is effectively learned by the ensemble models, with the main point of decision being where to "split" the product. As there are generally few locations where small molecules may be "split" effectively, these reaction subsets will tend toward higher performance, particularly as $k$ increases. Overall, reaction types that are learned well by the DiffER$^2$PG+ tend to have consistent and common reaction patterns with limited possible locations for a reaction center.

\subsection{Property Coefficient Analysis for DiffER$^2$ Models}
\label{app:property_coeffs}

\begin{figure}[h]
    \centering
    \includegraphics[width=0.95\linewidth]{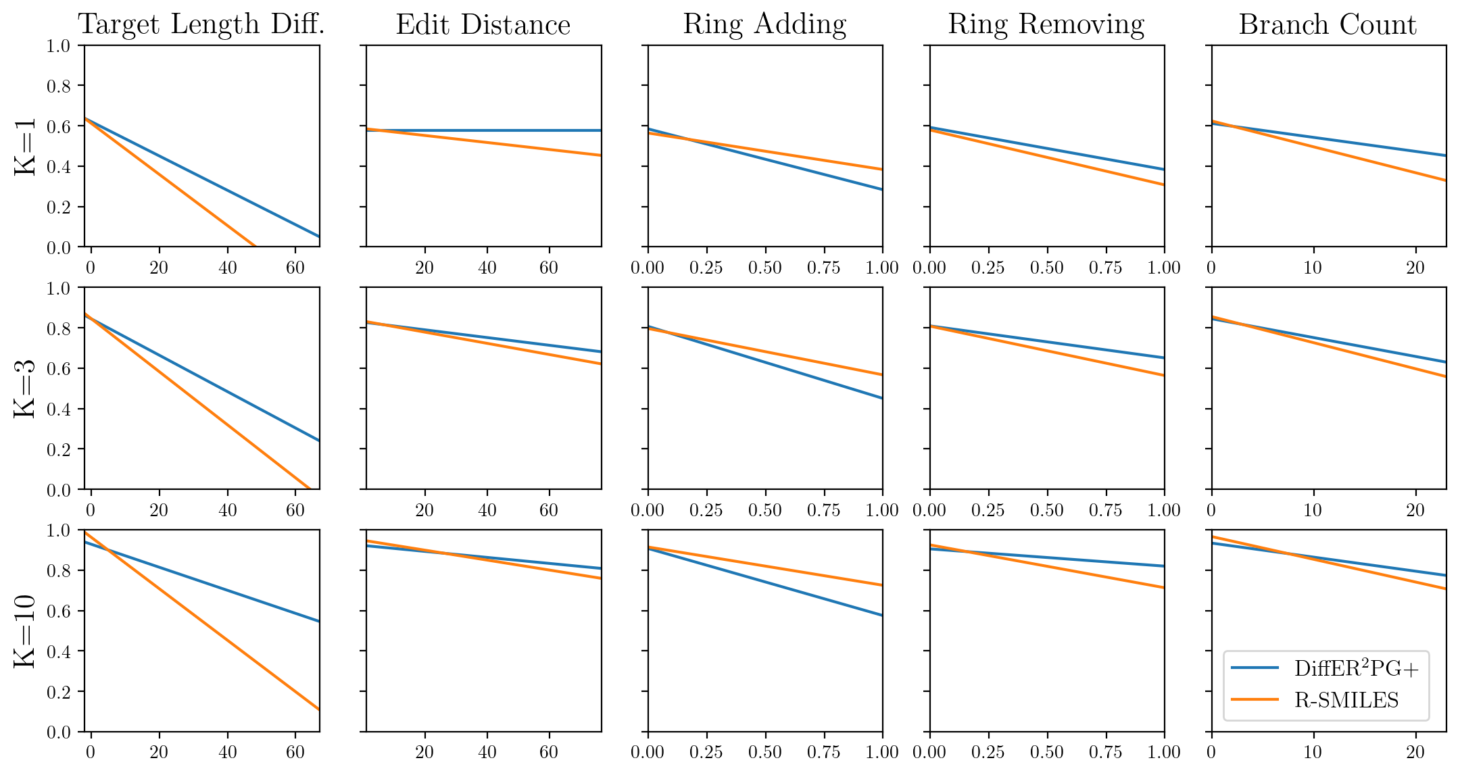}
    \caption{Lines of best fit between model performance and reactant properties for the DiffER$^2$PG+ and R-SMILES \cite{zhong2022root} models.}
    \label{fig:lines}
\end{figure}

In this appendix, we offer an analysis of the DiffER$^2$ models by regressing model performance with a variety of target reaction properties. These properties were selected due to the insight they give into the comparable and contrasting capabilities of diffusion and auto-regressive
models on SMILES reactions. These properties are:
\begin{enumerate}
    \item \textbf{Target Length Difference}: the difference in sequence length $L_x - L_y$ between the product SMILES and the target reactant SMILES. This property was chosen to characterize differing capabilities between diffusion models and auto-regressive models and can serve as a proxy for the leaving group size.
    \item \textbf{Edit Distance}: the Levenshtein distance between the product and reactant SMILES. This property characterizes how sequence similarity between the product and reactant affects performance.
    \item \textbf{Ring Adding}: If the number of rings in the product is greater than the number of rings in the reactant. This is indicative of ring formation. This represents a significant change in molecular and SMILES structure.
    \item \textbf{Ring Removing}: If the number of rings in the product is less than the number of rings in the reactant. Oftentimes, this is indicative of a large leaving group containing a ring structure rather than a true ring opening. This represents a significant change in molecular and SMILES structure.
    \item \textbf{Branch Count}: the number of branches in the target reactant SMILES. This is computed as the number of `(' tokens in the SMILES string. This represents structural complication in the SMILES string and molecular structure.
\end{enumerate}
 For each property, we report the slope of a line of best fit between the property statistic and the top-$k$ accuracy. A positive slope indicates that the model tends to perform well when the property statistic is high, and a negative slope indicates worse performance. In addition, we report the significance of each coefficient compared to the coefficients achieved by Root-aligned SMILES\cite{zhong2022root} to highlight significant differences between auto-regressive and diffusion models. This is accomplished by constructing the z-statistic $z = (\beta_{DiffER} - \beta_{RSmiles}) /\sqrt{SE_{DiffER}^2 + SE_{RSmiles}^2}$, where $\beta$ is the slope and $SE$ is the standard error \cite{d112b6d0-fe58-3096-80db-c75cfcf6946b}. These results are reported for the DiffER$^2$ ensemble and the continuous-time C-DiffER and C-DiffER$_{20}$ models in table \ref{tab:properties}, with a visualization of the best fit lines for each property provided in figure \ref{fig:lines}.

\begin{table}[t!]
    \small
    \centering
    \caption{Coefficients between reaction properties and performance metrics for the DiffER$^2$ ensemble and the C-DiffER model. The value in parenthesis indicates the change in coefficient compared to coefficients obtained using R-Smiles \cite{zhong2022root}. Values that are significantly different from $0$ are \underline{underlined} at the $95\%$ level and \textbf{bolded} at the $99\%$ level.}
    \begin{tabular}{l l c c c}
        \toprule
         Property & Model & K=1 & 3 & 10 \\
        \midrule
        \multirow{5}{*}{Target Length Diff.}    
            & C-DiffER &
            \textbf{-0.0092}~(\underline{+0.0035}) &
            \textbf{-0.0108}~(+0.0023) &
            \textbf{-0.0104}~(+0.0024) \\    
            & C-DiffER$_{20}$ &
            \textbf{-0.0071}~\textbf{(+0.0056)} &
            \textbf{-0.0084}~\textbf{(+0.0047)} &
            \textbf{-0.0072}~\textbf{(+0.0055)} \\
            & DiffER$^2$ & 
            \textbf{-0.0078}~\textbf{(+0.0049)} &
            \textbf{-0.0090}~\textbf{(+0.0041)} &
            \textbf{-0.0066}~\textbf{(+0.0061)} \\
            & DiffER$^2$PG & 
            \textbf{-0.0074}~\textbf{(+0.0052)} &
            \textbf{-0.0075}~\textbf{(+0.0056)} &
            \textbf{-0.0060}~\textbf{(+0.0067)} \\
            & DiffER$^2$PG+ & 
            \textbf{-0.0085}~(\underline{+0.0041}) &
            \textbf{-0.0090}~\textbf{(+0.0041)} &
            \textbf{-0.0057}~\textbf{(+0.0070)} \\
        \midrule
        \multirow{5}{*}{Edit Distance}    
            & C-DiffER &
            0.0003~(+0.0020) &
            \textbf{-0.0020}~(+0.0007) &
            \textbf{-0.0024}~(+0.0000) \\    
            & C-DiffER$_{20}$ &  
            0.0003~(\underline{+0.0021}) &
            \textbf{-0.0021}~(+0.0007) &
            \textbf{-0.0016}~(+0.0008) \\
            & DiffER$^2$ & 
            0.0000~(+0.0017) &
            \textbf{-0.0019}~(+0.0009) &
            \textbf{-0.0015}~(+0.0010) \\ 
            & DiffER$^2$PG & 
            0.0005~(\underline{+0.0022}) &
            \underline{-0.0014}~(+0.0014) &
            \textbf{-0.0016}~(+0.0009) \\
            & DiffER$^2$PG+ & 
            0.0000~(+0.0017) &
            \textbf{-0.0019}~(+0.0009) &
            \textbf{-0.0015}~(+0.0010) \\
        \midrule                             
        \multirow{5}{*}{Ring Adding}    
            & C-DiffER &
            \textbf{-0.2823}~(-0.1016) &
            \textbf{-0.2895}~(-0.0601) &
            \textbf{-0.3083}~(\underline{-0.1189}) \\    
            & C-DiffER$_{20}$ &
            \textbf{-0.2592}~(-0.0786) &
            \textbf{-0.3513}~(\underline{-0.1219}) &
            \textbf{-0.3400}~\textbf{(-0.1506)} \\
            & DiffER$^2$ & 
            \textbf{-0.2826}~(-0.1020) &
            \textbf{-0.3274}~(-0.0980) &
            \textbf{-0.3571}~\textbf{(-0.1677)} \\
            & DiffER$^2$PG & 
            \textbf{-0.2897}~(-0.1091) &
            \textbf{-0.3867}~\textbf{(-0.1574)} &
            \textbf{-0.3273}~\textbf{(-0.1379)} \\
            & DiffER$^2$PG+ & 
            \textbf{-0.3009}~(-0.1203) &
            \textbf{-0.3565}~(\underline{-0.1271}) &
            \textbf{-0.3320}~\textbf{(-0.1426)} \\
        \midrule                          
        \multirow{5}{*}{Ring Removing}    
            & C-DiffER &
            \textbf{-0.2316}~(+0.0403) &
            \textbf{-0.2397}~(+0.0048) &
            \textbf{-0.2258}~(-0.0136) \\    
            & C-DiffER$_{20}$ & 
            \textbf{-0.1891}~(\underline{+0.0828}) &
            \textbf{-0.1782}~(\underline{+0.0662}) &
            \textbf{-0.1656}~(+0.0466) \\
            & DiffER$^2$ & 
            \textbf{-0.2166}~(+0.0553) &
            \textbf{-0.1821}~(\underline{+0.0624}) &
            \textbf{-0.1397}~\textbf{(+0.0726)} \\   
            & DiffER$^2$PG & 
            \textbf{-0.1878}~(\underline{+0.0841}) &
            \textbf{-0.1453}~\textbf{(+0.0992)} &
            \textbf{-0.0957}~\textbf{(+0.1166)} \\
            & DiffER$^2$PG+ & 
            \textbf{-0.2088}~(+0.0631) &
            \textbf{-0.1585}~\textbf{(+0.0860)} &
            \textbf{-0.0854}~\textbf{(+0.1268)} \\
        \midrule
        \multirow{5}{*}{Branch Count}    
            & C-DiffER &
            \textbf{-0.0079}~(+0.0050) &
            \textbf{-0.0106}~(+0.0023) &
            \textbf{-0.0132}~(-0.0020) \\    
            & C-DiffER$_{20}$ &
            \textbf{-0.0081}~(+0.0048) &
            \textbf{-0.0106}~(+0.0024) &
            \textbf{-0.0098}~(+0.0015) \\
            & DiffER$^2$ & 
            \textbf{-0.0071}~(+0.0058) &
            \textbf{-0.0100}~(+0.0030) &
            \textbf{-0.0070}~(\underline{+0.0043}) \\  
            & DiffER$^2$PG & 
            \textbf{-0.0073}~(+0.0055) &
            \textbf{-0.0085}~(+0.0044) &
            \textbf{-0.0082}~(+0.0031) \\
            & DiffER$^2$PG+ & 
            \textbf{-0.0069}~(+0.0059) &
            \textbf{-0.0094}~(+0.0036) &
            \textbf{-0.0070}~(\underline{+0.0043}) \\
        \midrule
    \end{tabular}
    \label{tab:properties}
\end{table}

Table \ref{tab:properties} reports the slope coefficients, significance, and difference in coefficient values compared to R-SMILES for each property for each of the DiffER$^2$ models as well as the C-DiffER models. Figure \ref{fig:lines} offers a visualization of the lines of best fit for $k=1$, $k=3$, and $k=10$ performance conditioned on each of the analyzed properties. The results for $k=3$ and $k=10$ support the same conclusions discussed in the main text for $k=1$, primarily that:
\begin{enumerate}
    \item Diffusion models with length prediction are significantly more capable at predicting ground-truth reactions with larger increases in reactant SMILES length from the product SMILES compared to auto-regressive methods;
    \item Diffusion ensembles are significantly more capable at predicting ring-removing reactions compared to auto-regressive methods, but struggle to predict ring-adding reactions;
    \item Both diffusion models and auto-regressive models are negatively impacted by an increased edit distance between product and reactant SMILES as well as an increased branch count in the target reactant.
\end{enumerate}
Notably, however, many of the observed performance differences between R-SMILES and the DiffER$^2$ models become more significant at larger $k$, indicating that the trends observed at $k=1$ remain consistent across when more unique molecules are generated by the models. Furthermore, this highlights the capability for both model archetypes to predict common reaction rules that are likely to be used with high probability, resulting in $k=1$ distributions that perform more similarly than the more diverse $k=10$ molecule sets.

One interesting difference between DiffER$^2$PG+ and R-SMILES that becomes more significant at $k=10$ is the effect of branch count on model performance. While DiffER$^2$PG+ has higher slope for all values of $k$, the difference is only significant for $k=10$, as shown in Table \ref{tab:properties}. This difference in performance however is better observed in Figure \ref{fig:lines}, which also shows the difference in bias between DiffER$^2$PG+ and R-SMILES. Notably, when the branch count is low, R-SMILES performs slightly better than DiffER$^2$PG+, indicating the auto-regressive model may be stronger when the reactant SMILES has a highly sequential nature. However, as the number of branch terms in the reactant SMILES increases, the non-auto-regressive diffusion ensembles begin to exhibit stronger performance. This suggests that the diffusion models may be slightly more capable of generating SMILES sequences with non-sequential properties compared to state-of-the-art auto-regressive models.

\subsection{Case Study}
\label{app:case_study}

{
\renewcommand{\arraystretch}{1.1}
\begin{table}[h]
    \centering
    \caption{Case Studies. From left to right: the source product, the ground-truth reactant(s), and the top-3 predicted reactant(s) for the DiffER$^2$PG+ ensemble. Red highlighting in the source molecule indicates the ground truth change. Red highlighting in the true and predicted reactant(s) indicates differences between the source product and the reactants.}
    \label{tab:cases}
    \begin{tabular}{c}
         \toprule
         Case 1: Nitro Reduction to Aniline \\
         \includegraphics[width=0.95\linewidth]{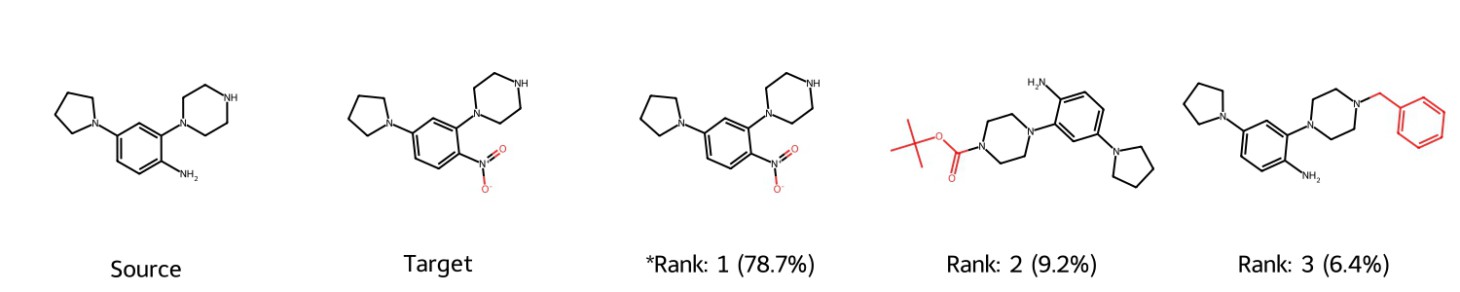} \\
         \midrule
         Case 2: Carbonyl Reduction to Alcohol\\
         \includegraphics[width=0.95\linewidth]{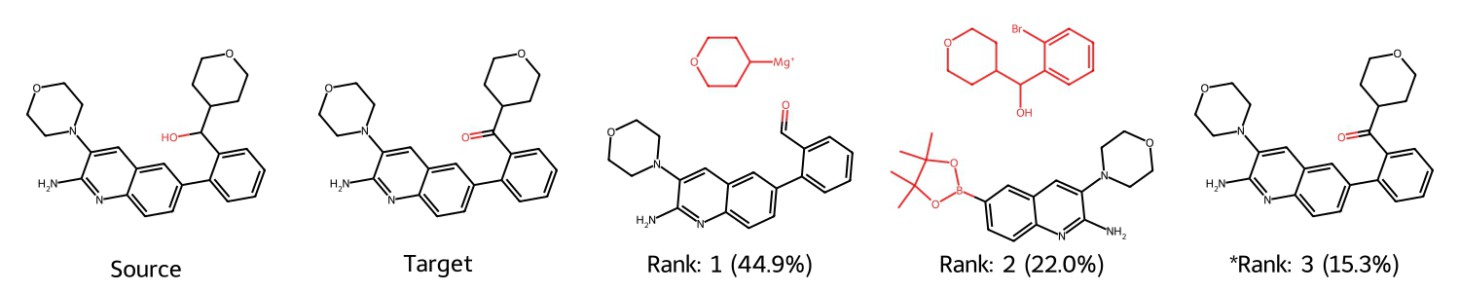} \\
         \midrule
         % Case 3: N-Heterocyclization with Dihalides\\
         % \includegraphics[width=0.95\linewidth]{images/case5.jpg} \\
         % \midrule
         Case 3: Deprotection using Methoxymethyl Ether\\
         \includegraphics[width=0.95\linewidth]{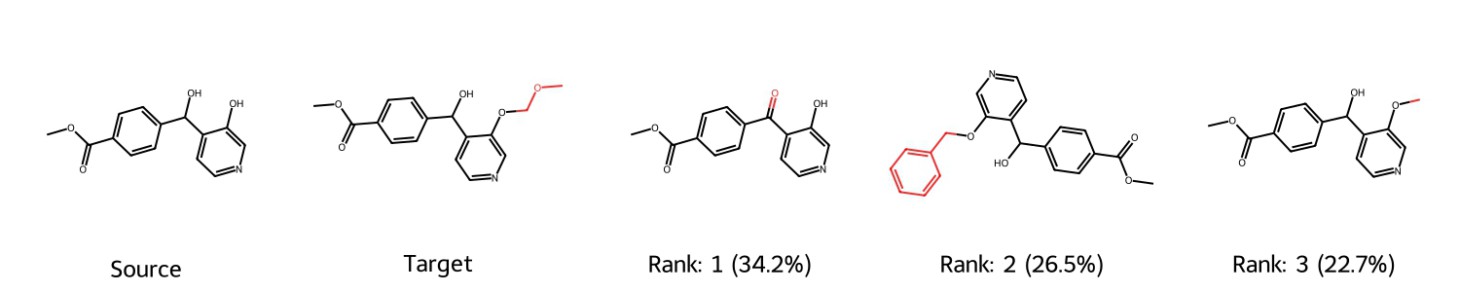} \\
         \midrule     
    \end{tabular}
\end{table}
}

In this section, we present three additional case studies on the DiffER$^2$PG+ ensemble. These reactions were selected to demonstrate the strengths and weakness of the ensemble. Table \ref{tab:cases} displays the source product, target reactant(s), and top-3 reactant(s) predicted by the DiffER$^2$PG+ ensemble.

\paragraph{Case 1: Nitro Reduction to Aniline.}
Case 1 presents a reduction of a nitro group (O=N-O$^-$) to a primary aromatic amine (NH$_2$). This is a common reaction that is generally achieved through metal catalyzed reduction of the nitro group, often using Pd, Zn, Fe, or Ni. This same reaction is the most commonly predicted reaction by the ensemble model, encompassing 78.7\% of all generated reactions. Instead of the nitro reduction, the second and third ranked reactions both take the form of an N-dealkylation to form a secondary amine. The rank 2 reaction is a proposed desterification in which the tert-butyl ester is reacted with a strong base such as NaOH to form an intermediate carboxylic acid and subsequently heated to decarboxylize, resulting in the target product. This is a viable alternative to the target reaction. The rank 3 reaction may be accomplished by Pt-catalyzed reaction with O$_2$ or by acid hydrolysis with hydrobromic or hydroiodic acid \cite{najmi2022n}. This reaction is also a viable alternative to the target reaction. Each of the top-3 reactions proposed by the ensemble is a viable way to synthesize the desired product. Reactions may be limited by the availability of the generated reactant(s).

\paragraph{Case 2: Carbonyl Reduction to Alcohol.}
Case 2 presents a common carbonyl reduction to an alcohol, in which a carbonyl group (C=O) is reacted with LiAlH$_4$ or NaBH$_4$ to form an alcohol (C-OH). This reaction is the third-most popular reaction produced by the ensemble model, appearing 15.3\% of the time. In contrast to the target reaction, The proposed rank 1 and rank 2 reactions add larger structural changes to core molecule, and may be reactions that could have been used during prior synthesis of the ground truth reactant. The rank 1 reaction is an aldehyde reaction with a Grignard reagent. This is a common and viable method to produce the product molecule. The rank 2 reaction is a proposed Suzuki cross-coupling, in which a boronic ester (R$_1$-B(OR)$_2$) is reacted with an organohalide (R$_2$-X) using a Palladium catalyst in the presence of a base to produce the target product (R$_1$-R$_2$). The Suzuki coupling may also be a viable alternative to the ground truth reaction. Like case 1, each of the top-3 reactants may be suitable methods to synthesize the desired product, limited primarily by the availability of reactants.

% \paragraph{Case 3: N-Heterocyclization with Dihalides.}
% Case 3 presents a N-hetercycle formation using dihalides, in which a primary amine is reacted with a dihalide in an alkaline aqueous medium to form a heterogeneous ring \cite{ju2006aqueous}. While this reaction is predicted by the ensemble model, it is only predicted 4\% of the time, making it the third most frequently predicted reaction. The rank 1 predicted reaction is a Borch reductive amination \cite{borch2003reductive} in which a carbonyl group is reacted with a secondary amine in the presence of a reducing agent to produce a tertiary amine. Unfortunately, due to the presence of multiple carbonyl groups in the predicted reactant, this reaction is likely to produce multiple unwanted byproducts. The rank 2 reaction is a substitution reaction with a mesylate ester and a secondary amide nucleophile. Again, this reaction does produce the desired product, but may produce still produce undesirable byproducts due to the electrophilicity of the carbamate group, which may also reaction with the amide nucleophile.

\paragraph{Case 3: Deprotection using Methoxymethyl Ether.}
Finally, case 3 presents a deprotection reaction in which a Methoxymethyl ether is removed via a silica-supported sodium hydrogen sulfate catalyst in dichloromethane solution \cite{ramesh2003simple}. Unfortunately, this reaction is not one of the top-3 reactions predicted by the ensemble, instead being produced at rank 7. The rank 1 prediction of the ensemble is a hydrolysis reaction in which a ketone is reduced to an alcohol, and is generally considered a common and trivial reaction. The rank 2 reaction is also a deprotection reaction involving a benzyl ether, which is removed using a metal catalyst in a hydrogen atmosphere. This is also a common and trivial reaction. Finally, the rank 3 predicted reaction is a acidic cleavage of a methyl ester. However, because this reaction is often done in an acidic environment, the presence of a ester group in the product/reactant may lead to unwanted byproducts. This reaction is non-trivial and is likely suboptimal compared to the rank 1, rank 2, and ground-truth reactions.

These case studies demonstrate both the effectiveness and pitfalls of the DiffER$^2$PG+ ensemble. As evidenced in all cases, the diffusion ensemble is capable of learning many common reaction patterns, but struggles to predict the ground-truth when many possible alternatives are present. While the reactions proposed by the DiffER$^2$ ensembles will generally produce the target, the model does not necessarily account for the availability of reactants or for the possible formation of side products, which may limit the models effectiveness. Furthermore, as evidenced by cases 2 and 3, it is not always clear what type of reaction is being desired: the model may be prone to larger structural changes (such as the top-2 reactants of case 2) over smaller changes that are necessary for preparing the molecule for future synthesis steps, such as protections, deprotections, and reductions. Moreover, the model may be biased toward reaction patterns that result in "easier" structural changes of the molecule, such as the top-2 reactions in the main case study (Figure \ref{fig:main_case}), in which the ensemble prefers reactants in which two structures are ``added'' together over the more ``advanced'' ring-forming reaction. Nevertheless, for each product in the case studies, DiffER$^2$PG+ does generate the target reactant(s) within the top-10 predictions, and in many cases provides viable alternate reactions. These case studies demonstrate that the diffusion ensemble is capable of learning effective reaction processes.

\subsection{Performance Analysis with Molecular Fingerprints}
\label{app:fingerprints}

Morgan fingerprints \cite{morgan1965generation} are a commonly used methodology for constructing feature vectors from molecule structures. The core of the method is built around substructure matching: each feature vector is a set number of bits, which each bit in the feature indicating whether or not a specific substructure is present in the molecule. These substructures capture the environment of atoms in a set radius and are generated using a hashing algorithm provided by the original authors \cite{morgan1965generation}. Because Morgan fingerprints capture many important characteristics of a molecule structure, we aim to investigate how the performance of the DiffER$^2$ models may be impacted by the presence of specific fingerprints in the target reactant(s), which could highlight specific substructures the ensembles may be biased against.  Additionally, we note that because Morgan fingerprints are constructed using a hashing function, Morgan bits are not necessarily unique indicators, as multiple substructures may map to a single index. However, they are still a useful tool for visualizing how performance relates to particular substructures.

We generate Morgan fingerprints with 2048 bits and a radius of 2 for the target reactants and analyze the correlation between the top-10 performance of the DiffER$^2$PG+ model and each bit in the generated fingerprint. We visualize the Morgan bits that have the most positive correlation with model performance as well as the Morgan bits that have the most negative correlation with model performance. All analyzed correlations are statistically significant at the $99\%$ ($p<0.01$) level. Visuals are produced by randomly selecting a reactant from the test set that has the associated fingerprint bit and visualizing the atom and neighborhood that produced the positive bit value. Atoms highlighted blue represent the central atom, while atoms highlighted yellow represent atoms in an aromatic ring. Atoms with gray highlights are in aliphatic rings.

\begin{figure}[h]
    \centering
    \includegraphics[width=0.9\linewidth]{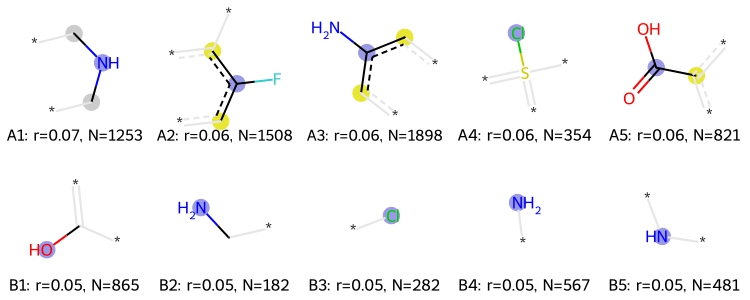}
    \caption{The 10 Morgan fingerprints with the highest positive correlation to Top-10 accuracy for the DiffER$^2$PG+ ensemble. "r" is the correlation value and "N" is the number of reactants in the test set that have the corresponding fingerprint. All correlations are statistically significant ($p < 0.01$).}
    \label{fig:top_10}
\end{figure}

Figure \ref{fig:top_10} visualizes the Morgan fingerprints that have the most positive correlation with model performance for $k=10$. These bits tend to be substructures with well-known and highly utilized reaction mechanisms and reactivity rules. For instance, fingerprint \textbf{A1} has the highest correlation with model performance, and represents an aliphatic secondary amine. Reactions with aliphatic secondary amines are well known and highly studied in organic synthesis, appearing in $\sim25\%$ of the test set. These reactions are commonly alyklation reactions, in which the amine is reacted with an alkyl halide to form new bonds and larger structures. Many fingerprints containing amines also exhibit positive correlation with model performance (\textbf{A3}, \textbf{B2}, \textbf{B4}, \textbf{B5}). Other notable fingerprints with positive correlation include: aryl halides (\textbf{A2}), which are commonly used in substitution reactions and as directing groups; carboxylic acids (\textbf{A5}) and alcohols (\textbf{B3}), two of the most studied functional groups in organic chemistry; halides (\textbf{B3}); and sulfonyl chlorides (\textbf{A4}), all of which are commonly used groups in organic synthesis.

\begin{figure}[b]
    \centering
    \includegraphics[width=0.9\linewidth]{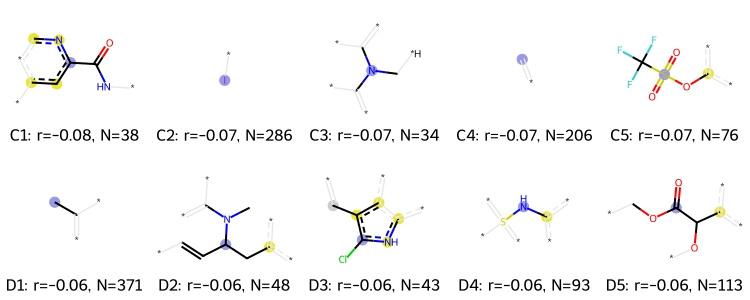}
    \caption{The 10 Morgan fingerprints with the highest negative correlation to Top-10 accuracy for the DiffER$^2$PG+ ensemble.  "r" is the correlation value and "N" is the number of reactants in the test set that have the corresponding fingerprint. All correlations are statistically significant ($p < 0.01$).}
    \label{fig:bottom_10}
\end{figure}

In contrast, fingerprints that have significant negative correlation with model performance tend to be rarer, have ambiguity in the reaction, or introduce complexity in the SMILES sequence structure. Figure \ref{fig:bottom_10} visualizes the Morgan fingerprints that have the most negative correlation with model performance for $k=10$. Many of these structures have a large number of branches or structural complexity (\textbf{C3}, \textbf{C5}, \textbf{D2}, \textbf{D3}, \textbf{D4}, \textbf{D5}), requiring SMILES structures with high non-sequential behavior. Many of these fingerprints may also require SMILES strings which detail stereochemistry of the molecule (\textbf{D1}, \textbf{D2}), further compounding the complexity of the SMILES string. In other cases, the fingerprint may require complexity in the SMILES string, but rather may represent ambiguity in the reactant. This is likely the case with \textbf{C2}, which represents a halide in the form of iodine. For many reactions, the explicit halide used may be interchangeable between Cl, Br, and I, as many chemical properties are shared between the three halogens with only subtle differences in reactivity and stability. A similar ambiguity may explain \textbf{C4}. Many reactions with alkenes result in the elimination of the C=C bond and the addition of new functional groups. To "reverse" this reaction, the retrosynthesis model would be have to identify a specific C-C bond as the reaction center. Because C-C bonds are extremely common, the model may have difficulty choosing these bonds as reaction centers, opting instead to target groups that are more commonly used in reactions. This behavior is readily observed in Appendix \ref{app:reaction_types}, Table \ref{tab:reaction_types}, which shows that the DiffER$^2$PG+ ensemble exhibits a reduced capability to predict reactions which form C-C bonds.

\subsection{Attention Map Analysis}
\label{app:attention}

Our final results are an analysis of the attention maps for a single continuous-time diffusion model with full sequence padding (C-DiffER). This model was used to better understand how the attention of the model changes throughout the diffusion process without the bias of a length-prediction component as described in Section \ref{sec:differ}. We provide visualizations of the attention maps for both the self attention and attention on the input product for each layer in the transformer model for $t=0.5$. Additionally, we map the attention on specific atoms in the SMILES sequences to the atoms they represent in the input and output molecular structures.

\begin{figure}[h!]
    \centering
    \includegraphics[width=\linewidth]{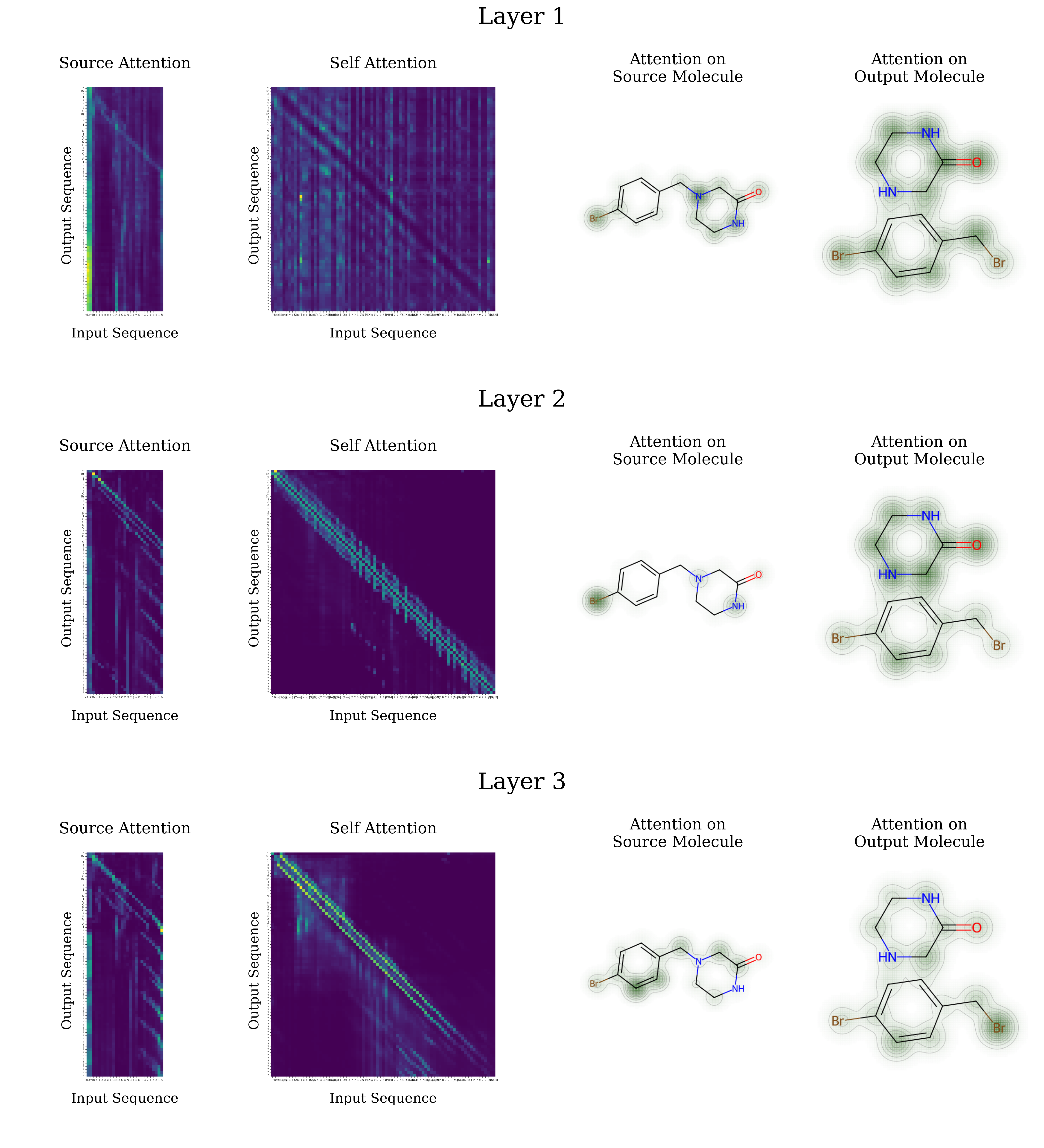}
    \caption{C-DiffER attention maps for layers 1, 2, and 3 on a reaction from the test set halfway through sampling, $t=0.5$. From left to right: attention on the input product SMILES $y_0$, attention on the noisy sample $x_t$, attention mapped to the input product molecule, and attention mapped to the final predicted output molecule.}
    \label{fig:attention_maps_0}
\end{figure}

\begin{figure}[h!]
    \centering
    \includegraphics[width=\linewidth]{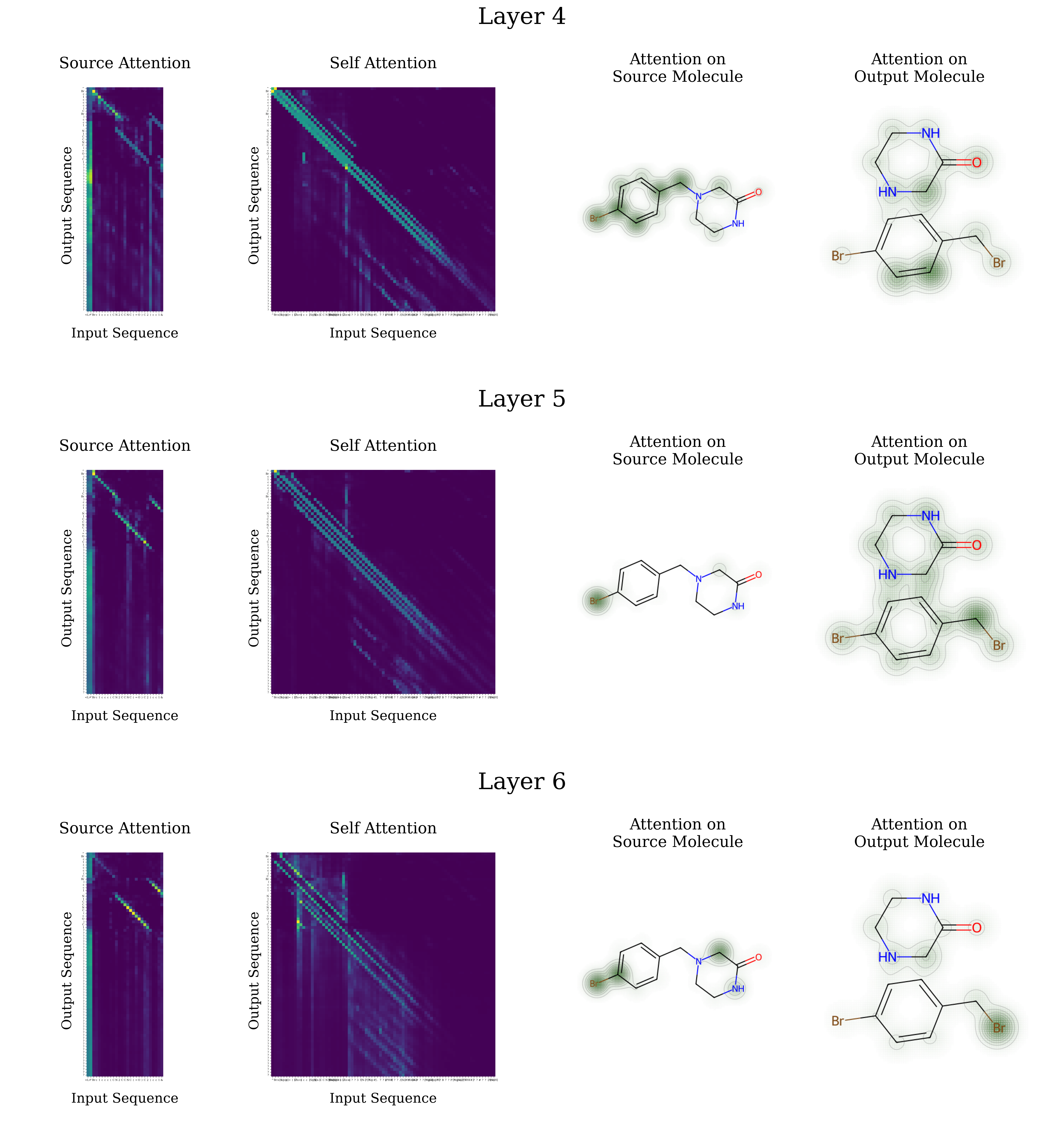}
    \caption{C-DiffER attention maps for layers 4, 5, and 6 on a reaction from the test set halfway through sampling, $t=0.5$. From left to right: attention on the input product SMILES $y_0$, attention on the noisy sample $x_t$, attention mapped to the input product molecule, and attention mapped to the final predicted output molecule.}
    \label{fig:attention_maps_1}
\end{figure}

Figures \ref{fig:attention_maps_0} and \ref{fig:attention_maps_1} show the attention maps for layers 1, 2, 3 and 4, 5, 6, respectively for a amine alkylation reaction. This reaction reacts an alkyl halide with an amine and is successfully produced as a top-1 reactant by the C-DiffER model. These attention maps demonstrate how the construction of the SMILES sequence differs from auto-regressive methods. Rather than building the sequence from the beginning, the attention maps suggest that the diffusion model works backward, attending to the full-length sequence initially and decreasing the attended sequence length as the model goes deeper. This helps explain why the diffusion models may have a stronger affinity for generating reactants with a larger change in sequence length over the product SMILES: unlike auto-regressive models, which may be likely to predict a STOP token earlier in the sequence, diffusion models attend to the maximum possible sequence at each step in the diffusion process, allowing for the full sequence to be predicted over during each estimate of $x_0$. As the iteration number increases, the diffusion model becomes more confident in its output of padding tokens at later timesteps, attending to those positions less after the first couple layers.

In the early self attention layers, the diffusion model appears to try and understand the general structure of the SMILES string, estimating the level of noise and confidence in the input $x_t$. This can be observed in the way sequence attention is spread across the entire sequence for each token rather than focused on tokens with similar locality. After the first layer, the self attention mechanism places much more emphasis on nearby positions, but still attends very little to each tokens' own position, indicating that the model is not confident in the fidelity of a one-to-one input mapping between $x_t$ and $x_s$. Interestingly, in the final self attention layer, significant attention is placed around the reaction centers in the generated reactants. Source attention layers tend to place attention near non-carbon atoms in the target product early on, expanding to carbon atoms only in the later layers. While this may simply be an indication that these atoms are "unique" in the molecule compared to the carbon atoms, they may also be used as "anchors" to build the generated reactant(s) around. This could also explain why the models struggle to predict C-C bond formation reactions, as it would require the model to identify a common C-C bond as a reaction center, which may not have any significant uniqueness compared to other structures in the product molecule. 

Another interesting feature of note is the repetitive attention patterns the predicted padding tokens place on the end of the product SMILES, notable in layers 2 and 3. Rather than the padding tokens placing attention on a single end token, the attention is stratified in a diagonal formation in groups of 10-12 tokens, attending primarily to the last 10-12 tokens of the product SMILES. This pattern is also observed in attention maps for other reactions, suggesting consistency in this behavior, which fades in the last couple layers. Repetition is a common phenomena also observed in large language models, which may repeat during inference time. The model only begins to reduce attention on later timesteps after this repetition begins, suggesting that it may use the repetition to "detect" padding locations. In the final layers of the transformer, attention on the source sequence is rather sparse and closely matches the attention maps observed by R-SMILES \cite{zhong2022root}, with the attention patterns "matching" tokens between the input and output SMILES.

The attention map analysis offers some insight into the workings of the diffusion models, and may help inform future model creation. While we offer some theories behind the meaning of these attention maps as it relations to SMILES generation for chemical retrosynthesis, significantly more investigation and experimentation needs to be done before our theories can be confirmed. We hope these attention maps provide some insight into the capabilities and methods of transformer diffusion models and promote further research into the explainability of transformer diffusion models. 

\newpage
